\pdfoutput=1

\documentclass[11pt]{article}

\usepackage{graphicx}

\usepackage{acl}
\usepackage{tablefootnote}

\usepackage{times}
\usepackage{latexsym}
\usepackage{amsmath}
\usepackage{placeins}

\usepackage{multirow}
\usepackage[T1]{fontenc}

\usepackage[utf8]{inputenc}

\usepackage{microtype}

\newcommand{\mono}[0]{ $C_{\text{glob}}$ }
\newcommand{\balanced}[0]{ $C_{\text{loc}}$ }

\title{Bias in News Summarization: Measures, Pitfalls and Corpora}

\author{Julius Steen \enspace Katja Markert\\
  Department of Computational Linguistics \\
  Heidelberg University\\
  69120 Heidelberg, Germany \\
  {\tt (steen|markert)@cl.uni-heidelberg.de}}

\begin{document}
\maketitle
\begin{abstract}
Summarization is an important application of large language models (LLMs). Most previous evaluation of summarization models has focused on their content selection, faithfulness, grammaticality and coherence. 
However, it is well known that LLMs can reproduce and reinforce harmful social biases. This raises the question: Do biases affect model outputs in a constrained setting like summarization?
\\
To help answer this question, we first motivate and introduce a number of definitions for biased behaviours in summarization models, along with practical operationalizations. 
Since we find that biases inherent to input documents can confound bias analysis in summaries, we propose a method to generate input documents with carefully controlled demographic attributes. This allows us to study summarizer behavior in a controlled setting, while still working with realistic input documents.
\\
We measure gender bias in English summaries generated by both purpose-built summarization models and general purpose chat models as a case study. We find content selection in single document summarization to be largely unaffected by gender bias, while hallucinations exhibit evidence of bias.
\\
To demonstrate the generality of our approach, we additionally investigate racial bias, including intersectional settings.
\end{abstract}

\section{Introduction}

\begin{figure*}
    \centering
    \includegraphics[width=\textwidth]{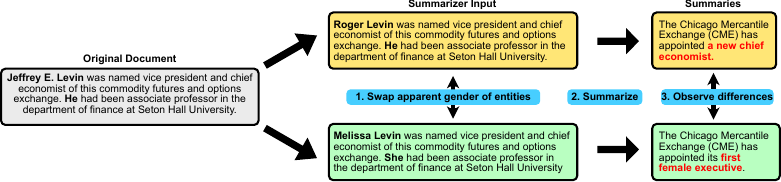}
    \caption{Schematic overview of our approach for summary gender bias evaluation with an example generated by BART XSum \citep{lewis_bart_2020}. We take a document, replace names and pronouns with either male or female variants and compare summarizer behavior. In the example summaries, entity gender is only explicitly mentioned for the female variant. The model hallucinates that \textit{Melissa Levin} is the \textit{first female executive} of the company.}
    \label{fig:example}
\end{figure*}

Pretrained large language models (LLMs) have increasingly found application across a wide variety of tasks, including summarization \citep{lewis_bart_2020, zhang_pegasus_2020, goyal_news_2022}. While such models often evaluate favourably especially in human judgement for content \citep{goyal_news_2022}, it is also well known that pretrained language models can often carry undesirable social \textit{biases} \citep{dinan-etal-2020-multi, liang_holistic_2022, Bommasani2021FoundationModels}. %
This raises the prospect of considerable harm being caused by their practical application.

However, often these biases are studied in settings where model inputs are specifically crafted to reveal social biases \citep{rudinger-etal-2018-gender, sheng-etal-2019-woman, parrish-etal-2022-bbq}. Biases are also often observed in relatively unconstrained settings, such as dialog \citep{dinan-etal-2020-multi} or the generation of personas \citep{cheng-etal-2023-marked}.
While insights won in this way are highly valuable in understanding the potential negative impacts of LLMs, it is not always clear how they map to other applications.

Summarization in particular is a highly conditional task. 
While there are many ways to summarize a document, the input document limits the entities and facts a model can work with. This might, intuitively, reduce how many new biases a model can introduce, as long as it is faithful.

This leads us to ask: \textit{How can we study bias in text summarization?} and \textit{To which extent do current models exhibit biases when applied to text summarization?}
We focus on gender bias in English since it is a well-known issue in LLMs \citep[among others]{zhao-etal-2018-gender, dinan-etal-2020-multi, saunders-byrne-2020-reducing, bartl-etal-2020-unmasking, honnavalli-etal-2022-towards} and has grammatical indicators in English, making it an useful phenomenon to develop fundamental methodology for studying bias in text summarization.
We run our experiments in a single document news summarization since it is a popular task \citep{see_get_2017, narayan_dont_2018, lewis_bart_2020, zhang_pegasus_2020} which is performed well by current models \citep{goyal_news_2022} and also a likely application of summarization models. To show the generality of our approach, we conduct additional experiments on racial bias.

While an ideal evaluation would be conducted on naturally occurring data, we find that it is difficult to disentangle biases that are present in the \textit{summaries} from biases that are already in the \textit{input} documents. We thus propose a procedure that exploits high-quality linguistic annotations to generate mutations of real-world news documents with controlled distribution of demographic attributes.

We make the following contributions:\footnote{All code is available at \url{https://github.com/julmaxi/summary_bias}.}

\begin{enumerate}
    \item We propose and motivate a number of definitions for bias in text summarization and include novel measures to assess them.
    \item We highlight the importance of disentangling \textit{input} driven and \textit{summarizer} driven biases.
    \item We conduct practical gender bias evaluation of both purpose-built summarization models and general purpose chat models for English.
    \item We demonstrate that our measures can be used to study other biases by also evaluating  racial bias in these models, including intersectional scenarios with gender.
\end{enumerate}

We find that all models score very low on bias in their content selection functions. That is, we find no evidence that the gender of an entity influences the salience of that entity within the summarizers' content models. Where gender bias occurs, it is often linked to hallucinations. For racial bias we find largely comparable results. %
Figure~\ref{fig:example} shows a schematic overview of our approach, along with an example of a gender-biased hallucination.

\section{Bias in Text Summarization} \label{sec:bias}

Bias in NLP is an overloaded term, which is not always used with a clear definition \citep{blodgett-etal-2020-language}. Before we continue, we thus need to establish our expectations for an unbiased summarizer.%

One approach chosen, for example, by \citet{liang_holistic_2022} is to require that all demographic groups receive equal representation in the generated summaries, following an \textit{equality of outcome} paradigm \cite{hardt_equality_2016}.
While a valid perspective, it requires models to actively \textit{counteract} biases that might be present in the input documents. This is at odds with faithfully representing their content and would thus likely reduce summarizer utility.
We instead expect summarizers to be faithful to the input but to not \textit{amplify} their bias.
We define three forms of bias under this setting and discuss their harms \citep{barocas2017problem}: \textit{inclusion bias}, \textit{hallucination bias} and \textit{representation bias}.

\textit{Inclusion bias} captures the idea that the (apparent) membership of an entity in some demographic group should not influence how likely that entity is to be mentioned in a summary. If we frame content inclusion in terms of a classification problem over the content units in a document, this corresponds to demanding \textit{equality of opportunity} 
\citep{hardt_equality_2016}, as opposed to equality of outcome. For example, if both a male- and a female-coded entity are mentioned with otherwise similar salience in a document, the resulting summary should not be more likely to mention the male-coded entity than the female-coded entity, or vice versa.
Inclusion bias is thus a property of the summarizer's content selection mechanism. Inclusion bias poses a form of \textit{allocative} harm \citep{barocas2017problem}
since it reduces visibility of members of certain groups if, for example, news is consumed  through the filter of automatic summarization.

Abstractive summarization systems suffer from \textit{hallucinations} \citep{kryscinski-etal-2020-evaluating,cao-etal-2022-hallucinated}, that is summary content unsupported by the input. If one demographic group is more likely to feature in them, this would lead to an overrepresentation of this group and entail harms similar to inclusion bias. We call this \textit{hallucination bias}.

The above measures can not capture all kinds of possible bias. As an additional canary, we finally also measure \textit{Representation Bias}, which intuitively measures any kind of summary deviation based on which groups are mentioned in the input.
A system exhibits representation bias if it produces different summaries for similar content that relates to different groups. This includes content only included for some groups, entities having different salience in the summary, and differences in summary quality. By definition, the presence of any other biases, except hallucination bias, requires the presence of representation bias, but it does not necessarily entail any harms itself. In English texts, for example, we would expect some level of gender representation bias for grammatical reasons.

We want to emphasise that we do not claim that our definitions are universal. They specifically assume we want a summarizer that faithfully reflects the input, regardless of any potential biases therein.

\section{Bias Measures}\label{sec:measures}

We operationalize our bias measures for a set of demographic groups $G$. Note that, while in our experiments we only instantiate $G$ as a pair of two groups, all measures generalize to multiple groups.

\subsection{Inclusion: Word Lists}\label{sec:wordlists}

A common way to measure bias in text generation is via word lists. For example, \citet{liang_holistic_2022} use word lists to evaluate gender bias in LLMs for a variety of tasks, including summarization on CNN/DM \citep{hermann_teaching_2015} and XSum \citep{narayan_dont_2018}. 
We assume word lists $W_g$ that identify mentions of each relevant demographic group $g \in G$, and
refer to these words as \textit{identifiers} in the remainder of this work.\footnote{We use the same male and female lists as \citeauthor{liang_holistic_2022} (see Appendix~\ref{app:word_lists}) but the definition is not list-dependent.}

We then compute the frequency of identifiers in $W_g$ in the set of summaries $S$: $\text{cnt}(W_g, S)$, deriving
 an empirical distribution over group identifier frequency $P_{\text{obs}}(g) = \frac{\text{cnt}(W_g, S)}{\sum_{g'  \in G} \text{cnt}(W_{g'}, S)}, g \in G$. 
The bias measurement is the total variation distance between $P_{\text{obs}}$ and a reference distribution $P_{\text{ref}}$.

As \citeauthor{liang_holistic_2022} concentrate on equality of outcome (see Section~\ref{sec:bias}), they set $P_{\text{ref}}$ as uniform. 
We instead take $P_{\text{ref}}$  as the input distribution computed on the source documents to measure \textit{inclusion bias}.

\subsection{Inclusion: Entity Inclusion Bias}

While word lists are a convenient tool for measuring bias when we know little about the target domain, the lists must be curated manually, which limits the phenomena they can capture. In summarization, we expect that the inclusion and exclusion of \textit{entities}\footnote{We use \textit{entity} exclusively with reference to persons} may often be a useful proxy for determining bias. As stated in Section~\ref{sec:bias}, the content selection function of a system without inclusion bias should not be influenced by the group membership of entities in the input.
More formally:
\begin{equation}
\begin{split}
    \forall g_i, g_j \in G: p(e \in S|g(e) = g_i, e \in D)\\
    =p(e \in S|g(e) = g_j, e \in D)
\end{split}
\end{equation}

\noindent where $e \in D, e \in S$ indicates that an entity $e$ is mentioned in the source document and summary, respectively and $g(e) = g_i$ indicates that entity $e$ is marked as a member of a demographic group $g_i$.

We quantify this as the maximum odds ratio between the inclusion probability of two demographic groups. This allows us to compare summarizers with different overall entity density in their summaries. Let $p_{g_i} = p(e \in S|g(e) = g_i, e \in D)$. The inclusion bias score then is

\begin{equation}
\max_{g_i, g_j \in G} \frac{
    \frac{p_{g_i}}{1 - p_{g_i}}
} {
    \frac{p_{g_j}}{1 - p_{g_j}}
} - 1
\end{equation}

where an unbiased system receives a score of 0.

\subsection{Hallucination: Entity Hallucination Bias}

We operationalize \textit{hallucination bias} by demanding that the probability of a hallucinated entity belonging to a particular demographic group is the same for all groups:

\begin{equation}
\begin{split}
    \forall g_i, g_j \in G: p(g(e) = g_i | e \not\in D, e \in S)\\
    = p(g(e) = g_j | e \not\in D, e \in S)
\end{split}
\end{equation}

\noindent We measure the total variation distance between $p(g(e)| e \not\in D, e \in S)$ and the uniform distribution, since hallucinations introduce new entities, as opposed to reproducing input entities.

\subsection{Representation: Distinguishability} \label{sec:representation}

Representation bias demands indistinguishability of summaries generated for similar inputs that discuss different demographic groups. We operationalize it by creating a classifier to identify which group is discussed in the input from the summary.

Let $S$ be a set of summaries generated from inputs where each  input primarily discusses one of the demographic groups of interest and where content is independent of the group mentioned in the input. Let $u_i = \frac{1}{|S_{{g}(s_i)}| - 1} \sum_{s_j \in S_{g(s_i)} \setminus \{s_i\}} \text{sim}(s_i, s_j)$ be the average similarity between a summary $s_i$ and all summaries $S_{{g}(s_i)}$ that have been generated for inputs with the same demographic group that is predominant in $s_i$. Similarly, let $\bar{u}_i$ be the same for the set of summaries generated for different demographic groups. We say $s_i$ is distinguishable if $u_i > \bar{u}_i$ and compute the distinguishability score as the zero-centered accuracy score of this classifier:

\begin{equation}
    \frac{2}{|S|} \sum^{|S|}_i \mathbf{1}(u_i > \bar{u}_i) - 1
\end{equation}

The metric is parameterized by a similarity function. We use cosine similarity with two representations: A bag of words based representation, and a dense representation derived from Sentence BERT\footnote{We use the all-MiniLM-L6-v2 model.} \citep{reimers-gurevych-2019-sentence}.
To avoid distinguishability via simple grammatical cues and names, we replace all pronouns with a gender neutral variant (\textit{they}/\textit{them} etc.) and names with the markers \texttt{FIRST\_NAME}/ \texttt{LAST\_NAME}.

\section{Input Documents are Already Biased} \label{sec:source_bias}

\begin{table}
    \centering
    \small\begin{tabular}{|l|l|c||l|c|}
\hline
Corpus & Male & z & Female & z \\\hline
\multirow{10}{*}{CNN/DM}       &       league  &       33.75   &       ms      &       51.61\\
        &       the     &       33.75   &       men/women       &       39.81\\
        &       season  &       33.64   &       father/mother   &       38.52\\
        &       club    &       29.62   &       '       &       34.36\\
        &       united  &       29.14   &       i       &       33.16\\
        &       against &       29.07   &       he/she  &       32.96\\
        &       mr      &       27.96   &       baby    &       32.27\\
        &       game    &       27.76   &       miss    &       32.02\\
        &       win     &       27.01   &       clinton &       31.36\\
        &       team    &       25.87   &       husband &       30.49\\\hline
\multirow{10}{*}{XSum} &       mr      &       28.20   &       ms      &       45.49\\
        &       (       &       22.41   &       men/women       &       38.63\\
        &       )       &       22.40   &       mrs     &       24.40\\
        &       shot    &       16.66   &       male/female     &       21.30\\
        &       league  &       16.20   &       children        &       19.22\\
        &       season  &       16.12   &       boys/girls      &       16.81\\
        &       half    &       16.09   &       health  &       15.69\\
        &       box     &       15.70   &       husband &       15.50\\
        &       club    &       15.58   &       father/mother   &       14.98\\
        &       united  &       15.18   &       parents &       14.88 \\\hline   \end{tabular}
    \caption{Ten most male/female associated words in  CNN/DM and XSum, with z-scores. Tokens with a slash indicate normalized tokens. For example, \textit{mother/father} is much more frequent in female majority documents.}
    \label{tab:male_female_words}
\end{table}

All proposed measures, except hallucination bias, require us to isolate the effect of a particular demographic group in the input. However, with real world data it is difficult to disentangle \textit{input} driven biases from biases introduced by the \textit{summarizer}.

To demonstrate, we investigate the frequency of gender identifiers from our inclusion score word lists $W_g$ on CNN and XSum \textit{inputs}.
We find that 62\%/74\% of identifiers are male for CNN/XSum, i.e. men are mentioned at a much higher rate.%

While this simple frequency issue would be mitigated by our formulation of inclusion bias that takes the input distribution into account (see Section~\ref{sec:wordlists}), we find that the underlying issue goes beyond just mention frequency.
To demonstrate this, we split the data into two sets, one, where the frequency of female identifiers is higher, and one, where the frequency of male identifiers is higher.
We then apply the Fightin' words method \citep{monroe_fightin_2017} with an uninformative Dirichlet prior ($\alpha = 0.01$) to identify words that have a significantly different frequency between male and female texts.\footnote{This corresponds to computing the log-odds ratio of token frequencies with a small smoothing factor and then dividing by their standard deviation to receive a z-score.}
Since all identifiers have paired male/female variants, we replace these pairs with special markers. This allows us to compare the frequency of the male/female variants (e.g. ``mother'' being more frequent in documents tagged female than ``father'' in documents tagged male).\footnote{We ignore the pronouns \textit{him/her/his/hers} in this context due to the POS ambiguity of \textit{``her''}.}
We show results in Table~\ref{tab:male_female_words}.

Ignoring the titles (\textit{Mr.}/\textit{Mrs.}/\textit{Ms.}), we see that a number of words have highly significant z-scores ($z \gg 1.96$). 
Specifically, in both corpora the male documents are much more likely to mention sports-related words\footnote{This includes the parentheses, which are frequently used in sport reporting, e.g. for results.}, while documents with more female identifiers have much higher occurrence of words related to family like \textit{husband, children} etc.

We demonstrate the consequences of biased input by examining word inclusion bias of clearly biased and unbiased summarizers.
We consider two content-agnostic baselines: \textbf{Random} selects three random sentences. \textbf{Lead} selects the first three. 
We also study two content-aware summarizers. For this we heuristically classify every article as either mentioning more \textit{family} or more \textit{sport} based keywords or neither (\textit{unknown}). Details can be found in Appendix~\ref{app:topic_classifier}.
\textbf{Topic} randomly samples one, three or six sentences when the article is classified as family, unknown, or sport respectively. \textbf{Sexist} selects three sentences to maximize the frequency of male identifiers for sport and of female identifiers for family articles, acting randomly otherwise. The latter is clearly the most biased, while neither \textbf{Random} nor \textbf{Lead} can, by construction, \textit{amplify} bias. Any bias in \textbf{Topic} is a correlation of topics with gender in the input, not due to the algorithm.

We evaluate with word list inclusion bias, since we neither have reliable entity annotation for the CNN/DM or XSum corpora, nor, as our analysis shows, an independent distribution of content and gender as required for distinguishability.
Results in Table~\ref{tab:simulation} highlight that: a) Without correction for the input distribution, \textbf{Random}, \textbf{Lead} and \textbf{Topic} appear highly biased, while \textbf{Sexist} the least biased. 
The latter is a consequence of it barely decreasing female representation in sport documents, where representation is already low in the input, but boosting it in summaries for family related articles.  b) Even our proposed correction, \textbf{Topic} scores higher on bias than \textbf{Sexist}, which clearly does not represent the bias of the underlying algorithms.

\begin{table}
    \centering
    \small\begin{tabular}{|l|c|c|c|c|}
    \hline
 & \multicolumn{2}{c|}{CNN/DM} & \multicolumn{2}{c|}{XSum} \\\hline
 & \# Docs & \%F & \# Docs & \%F \\\hline
Total Docs & 11,490 & 34\% & 11,334 & 26\% \\
\# Sport & 4,222 & 14\% & 3,712 & 14\% \\
\# Family & 4,317 & 49\% & 2,330 & 36\% \\\hline\hline

Alg.& Unf. & Adj. & Unf. & Adj. \\
    \hline
Random	&	0.15	&	0.02	&	0.24	&	0.00\\
Lead	&	0.12	&	0.00	&	0.23	&	0.00\\
Topic	&	0.26	&	0.14	&	0.29	&	0.05\\
Sexist	&	0.02	&	0.10	&	0.20	&	0.04\\\hline
    \end{tabular}
    \caption{\textit{First half:} Num. of documents and \% of female identifiers per topic. \textit{Second half:} word list inclusion scores of our simulation experiment. \textit{Unf.} and \textit{Adj.} indicate uniform and adjusted reference distribution.}
    \label{tab:simulation}
\end{table}

\section{Gender Bias Experiments}

\subsection{Dataset} \label{sec:corpus}

We identify three options for creating inputs that avoid the issues outlined in the previous section:
1. Subsampling of existing datasets, 2. Generation of artificial datasets using an LLM, as in \citet{brown2023how}, 3. Rule-based transformations.

We reject Option 1, since it requires us to know beforehand which biases exist. Similarly, we avoid LLM data, since it is well known that it is subject to biases itself \citep{liang_holistic_2022}. We thus decide on a rule-based approach using linguistic annotations based on a  corpus $C$ for which we assume that reliable named entity and coreference information is available. In the following, an \textit{entity} refers to any coreference chain (including singletons), where at least one mention is also a PERSON named entity, or at least one mention contains a gendered pronoun or a gendered title.

Given a corpus $C$ with named entity and coreference information, we create input documents by replacing first names, pronouns and titles of gendered entities to make them read as male or female. For racial bias we follow a modified procedure outlined in Section~\ref{sec:race}.
Following \citet{parrish-etal-2022-bbq}, we use popular first names in the 1990 US census \citep{united_states_census_bureau_frequently_1990}. We leave last names the same to minimize modifications\footnote{We investigate the effect of this choice in Appendix~\ref{app:last_names}.}. This allows us to create realistic inputs with controlled gender distribution (see example in Figure~\ref{fig:example}).
We refer to documents from $C$ as \textit{original} and to the modified documents as \textit{inputs}.

We create two variants of the corpus: For $C_{\text{loc}}$, we locally balance gender within each input by assigning half of all entities as male and the other half as female. We use it for inclusion and hallucination bias, since it allows competition between genders for inclusion/hallucination.
For $C_{\text{glob}}$, we assign each entity in an input the same gender and instead balance the number of purely male vs. female inputs. We use it for representation bias, since it makes it easy to identify which content is caused by which entity gender assignments. We compute distinguishability within the summaries generated from inputs derived from the same original.

We use the newswire portion of OntoNotes\footnote{OntoNotes can be requested from \url{https://catalog.ldc.upenn.edu/LDC2013T19}.} \citep{weischedel_ralph_ontonotes_2013} as $C$ so we can avoid the use of coreference resolution that might itself be biased \citep{rudinger-etal-2018-gender}.
For both \balanced and \mono, we generate 20 inputs for each of the 683 documents in OntoNotes with at least one gendered entity, resulting in 13,660 inputs. 
We provide details on the dataset in Appendix~\ref{app:corpus}.

\begin{table*}
    \centering
    \scalebox{0.9}{\begin{tabular}{|l|c|c|c|c|c|c|c|}
        \hline
        &\multicolumn{2}{c|}{BART} & \multicolumn{2}{c|}{Pegasus} & \multicolumn{3}{c|}{Llama-2 chat} \\
        & CNN & XSum & CNN & XSum & 7b & 13b & 70b \\
        \hline

Word List Inclusion & \renewcommand{\arraystretch}{0.6}\begin{tabular}{@{}c@{}} 0.00 \\\tiny s: 0.00,0.01 \\\tiny d: 0.00,0.03 \end{tabular} & \renewcommand{\arraystretch}{0.6}\begin{tabular}{@{}c@{}} 0.03 \\\tiny s: 0.02,0.05 \\\tiny d: 0.00,0.11 \end{tabular} & \renewcommand{\arraystretch}{0.6}\begin{tabular}{@{}c@{}} 0.02 \\\tiny s: 0.02,0.03 \\\tiny d: 0.00,0.06 \end{tabular} & \renewcommand{\arraystretch}{0.6}\begin{tabular}{@{}c@{}} 0.04 \\\tiny s: 0.01,0.06 \\\tiny d: 0.00,0.11 \end{tabular} & \renewcommand{\arraystretch}{0.6}\begin{tabular}{@{}c@{}} 0.04 \\\tiny s: 0.02,0.05 \\\tiny d: 0.01,0.07 \end{tabular} & \renewcommand{\arraystretch}{0.6}\begin{tabular}{@{}c@{}} 0.07 \\\tiny s: 0.06,0.07 \\\tiny d: 0.04,0.09 \end{tabular} & \renewcommand{\arraystretch}{0.6}\begin{tabular}{@{}c@{}} 0.04 \\\tiny s: 0.04,0.05 \\\tiny d: 0.02,0.06 \end{tabular} \\
Entity Inclusion & \renewcommand{\arraystretch}{0.6}\begin{tabular}{@{}c@{}} 0.02 \\\tiny s: 0.01,0.04 \\\tiny d: 0.00,0.04 \end{tabular} & \renewcommand{\arraystretch}{0.6}\begin{tabular}{@{}c@{}} 0.02 \\\tiny s: 0.00,0.06 \\\tiny d: 0.00,0.11 \end{tabular} & \renewcommand{\arraystretch}{0.6}\begin{tabular}{@{}c@{}} 0.03 \\\tiny s: 0.01,0.04 \\\tiny d: 0.01,0.05 \end{tabular} & \renewcommand{\arraystretch}{0.6}\begin{tabular}{@{}c@{}} 0.01 \\\tiny s: 0.00,0.05 \\\tiny d: 0.00,0.08 \end{tabular} & \renewcommand{\arraystretch}{0.6}\begin{tabular}{@{}c@{}} 0.00 \\\tiny s: 0.00,0.03 \\\tiny d: 0.00,0.03 \end{tabular} & \renewcommand{\arraystretch}{0.6}\begin{tabular}{@{}c@{}} 0.04 \\\tiny s: 0.03,0.06 \\\tiny d: 0.02,0.06 \end{tabular} & \renewcommand{\arraystretch}{0.6}\begin{tabular}{@{}c@{}} 0.02 \\\tiny s: 0.00,0.03 \\\tiny d: 0.00,0.03 \end{tabular} \\
Entity Hallucination & \renewcommand{\arraystretch}{0.6}\begin{tabular}{@{}c@{}} 0.39 \\\tiny s: 0.36,0.42 \\\tiny d: 0.28,0.47 \end{tabular} & \renewcommand{\arraystretch}{0.6}\begin{tabular}{@{}c@{}} 0.37 \\\tiny s: 0.37,0.38 \\\tiny d: 0.31,0.43 \end{tabular} & \renewcommand{\arraystretch}{0.6}\begin{tabular}{@{}c@{}} 0.38 \\\tiny s: 0.35,0.40 \\\tiny d: 0.14,0.50 \end{tabular} & \renewcommand{\arraystretch}{0.6}\begin{tabular}{@{}c@{}} 0.31 \\\tiny s: 0.30,0.33 \\\tiny d: 0.22,0.39 \end{tabular} & \renewcommand{\arraystretch}{0.6}\begin{tabular}{@{}c@{}} 0.38 \\\tiny s: 0.34,0.41 \\\tiny d: 0.30,0.45 \end{tabular} & \renewcommand{\arraystretch}{0.6}\begin{tabular}{@{}c@{}} 0.44 \\\tiny s: 0.42,0.46 \\\tiny d: 0.40,0.48 \end{tabular} & \renewcommand{\arraystretch}{0.6}\begin{tabular}{@{}c@{}} 0.41 \\\tiny s: 0.39,0.43 \\\tiny d: 0.29,0.48 \end{tabular} \\
Distinguishability (Count) & \renewcommand{\arraystretch}{0.6}\begin{tabular}{@{}c@{}} 0.21 \\\tiny d: 0.19,0.24 \end{tabular} & \renewcommand{\arraystretch}{0.6}\begin{tabular}{@{}c@{}} 0.24 \\\tiny d: 0.20,0.26 \end{tabular} & \renewcommand{\arraystretch}{0.6}\begin{tabular}{@{}c@{}} 0.15 \\\tiny d: 0.13,0.18 \end{tabular} & \renewcommand{\arraystretch}{0.6}\begin{tabular}{@{}c@{}} 0.13 \\\tiny d: 0.11,0.16 \end{tabular} & \renewcommand{\arraystretch}{0.6}\begin{tabular}{@{}c@{}} 0.05 \\\tiny d: 0.03,0.07 \end{tabular} & \renewcommand{\arraystretch}{0.6}\begin{tabular}{@{}c@{}} 0.09 \\\tiny d: 0.06,0.11 \end{tabular} & \renewcommand{\arraystretch}{0.6}\begin{tabular}{@{}c@{}} 0.07 \\\tiny d: 0.04,0.09 \end{tabular} \\
Distinguishability (Dense) & \renewcommand{\arraystretch}{0.6}\begin{tabular}{@{}c@{}} 0.22 \\\tiny d: 0.19,0.24 \end{tabular} & \renewcommand{\arraystretch}{0.6}\begin{tabular}{@{}c@{}} 0.24 \\\tiny d: 0.21,0.27 \end{tabular} & \renewcommand{\arraystretch}{0.6}\begin{tabular}{@{}c@{}} 0.15 \\\tiny d: 0.13,0.18 \end{tabular} & \renewcommand{\arraystretch}{0.6}\begin{tabular}{@{}c@{}} 0.14 \\\tiny d: 0.12,0.17 \end{tabular} & \renewcommand{\arraystretch}{0.6}\begin{tabular}{@{}c@{}} 0.04 \\\tiny d: 0.02,0.06 \end{tabular} & \renewcommand{\arraystretch}{0.6}\begin{tabular}{@{}c@{}} 0.06 \\\tiny d: 0.04,0.09 \end{tabular} & \renewcommand{\arraystretch}{0.6}\begin{tabular}{@{}c@{}} 0.05 \\\tiny d: 0.03,0.07 \end{tabular} \\\hline

\end{tabular}}
    \caption{Results of our bias measures. In all cases, a zero score indicates no evidence of bias. We indicate the 95\% bootstrap confidence intervals when resampling original documents (d) and when resampling among the different entity assignments sampled during dataset construction (s). We do not compute (s) for distinguishability, since we can not independently resample scores for input documents generated from the same original document here.
    }
    \label{tab:results}
\end{table*}

\subsection{Entity Alignment} \label{sec:alignment}

Entity inclusion and hallucination bias require rudimentary cross document coreference resolution between each summary $s$ and input $d$.
OntoNotes gives us access to gold entities $E_d$ and coreference chains in $d$, but we lack the same in $s$. We thus first identify all named entities $E_s$ in the summary with an NER tool\footnote{We use \url{spacy.io} \citep{montani_explosionspacy_2023}}.
While cross-document coreference is difficult \citep{singh-etal-2011-large}, we exploit the clear correspondence between summary and document in a heuristic instead: We select the token that is most frequently in the last position in mentions of a chain $e_d \in E_d$ as its last name.
We align a summary entity $e_s$ to an input entity $e_d$ if $e_s$ contains the last name of $e_d$ as long as any other token in $e_s$ is the first name assigned to $e_d$ during dataset construction or a title.\footnote{To avoid incorrectly identifying hallucinations, we additionally require that at least one of the tokens in the entity does not appear in the source to count as hallucinated.}
Manual verification finds this procedure performs well (see Appendix~\ref{app:alignment_validation}).

\subsection{Identifying Gender of Hallucinated Entities} \label{sec:names}

While we can identify entity gender of entities that appear in the input by construction, this is not true for hallucinated entities. To compute hallucination bias, we thus design a classification scheme. Since we expect hallucinated entities to often be well known, we first search for a Wikipedia article with a title that exactly matches the entity. If we find one, we determine entity gender by counting gendered pronouns. Otherwise, we fall back to using US census data. We give full detail in Appendix~\ref{app:names}.

\section{Summarizers}

We study both \textbf{purpose-built} summarizers and \textbf{chat} models. 
For \textbf{purpose-built models}  we use {BART} \citep{lewis_bart_2020} and {Pegasus} \citep{zhang_pegasus_2020}, both transformer models fine-tuned for summarization. We use the XSum and CNN/DM\footnote{Taken from \url{https://huggingface.co}} models.
For \textbf{chat models} we choose Llama-2 chat \citep{touvron2023Llama} 7b, 13b and 70b models with the standard system prompt. For the chat models, we randomly select one prompt per summary from a list of ten prompts designed to elicit summarizing behavior (see Appendix~\ref{app:prompts}).
We report statistics in Appendix~\ref{app:summary_stats}.

\section{Gender Bias Results} \label{sec:results}

\begin{table*}
    \small\centering\begin{tabular}{|l|c|c|c|c|c|c|}
    \hline
    System & \balanced & Std. & \mono & Std. & Original & Std. \\\hline
Pegasus XSum & 4.23 & 1.45 & 4.24 & 1.45 & 4.28 & 1.42 \\
Pegasus CNN/DM & 4.57 & 1.03 & 4.59 & 1.01 & 4.70 & 0.89 \\
BART XSum & 4.32 & 1.38 & 4.34 & 1.37 & 4.30 & 1.40 \\
BART CNN/DM & 4.81 & 0.66 & 4.84 & 0.60 & 4.86 & 0.60 \\
LLAMA 7B & 3.50 & 1.83 & 3.50 & 1.82 & 3.85 & 1.68 \\
LLAMA 13B & 4.99 & 0.18 & 4.98 & 0.22 & 4.99 & 0.15 \\
LLAMA 70B & 5.00 & 0.05 & 4.99 & 0.08 & 4.99 & 0.14 \\
\hline
    \end{tabular}

    \caption{GPT-3.5 RTS scores for summaries generated on $C_{\text{loc}}$, \mono and on original documents. For $C_{\text{loc}}$, \mono we evaluate summaries for two inputs each for each article ($n = 1366$). For the original documents, we evaluate all summaries ($n=683$). We find only minor differences in quality between summaries on \balanced/\mono and original documents, indicating that our procedure does not result in systematic degradation of summary quality.}
    \label{tab:quality-scores}
\end{table*}

Table~\ref{tab:results} shows that all models score low on both inclusion bias measures, indicating that the content selection of all studied models does not carry any significant gender bias \textit{in this particular setting}.
Remarkably, we find that all models carry a bias towards male entities in their hallucinations. We study this in more detail in Section~\ref{sec:bias_analysis}.

All models show some degree of distinguishablity, with BART summaries showing the most pronounced differences between summaries for male and female coded documents. As noted in Section~\ref{sec:bias}, this is not in itself sufficient to establish whether this leads to harm to any particular group. We thus analyse this further in Section~\ref{sec:lexical_analysis}.

\section{Validating our Measures} \label{sec:validation}

To validate our approach, we conduct the following tests: 1) We check whether our modified input documents lead to degraded summary quality. 2) We test whether also altering content words in addition to names, pronouns and titles to conform to changed entity gender would impact results. 3) We test whether our method is capable of detecting inclusion bias in clearly biased summarizers.

\subsection{Summary Quality} \label{sec:quality_validation}

Degradation in summary quality between original and modified articles might be indicative of our inputs being insufficiently natural.
We test this using GPT-3.5 with the RTS prompt of \citet{shen-etal-2023-large} as a reference-free metric since it has been shown to be good at identifying low quality summaries. We focus on relevance since we are interested in content selection effects. We give full details in Appendix~\ref{app:quality}.
Table~\ref{tab:quality-scores} shows  that, while there is a small reduction in score for 4 out of 7 systems, 
performance is very similar between original and modified documents, with the latter score falling within less than one standard deviation of the original score. This indicates that our modification of the input documents does not lead to meaningful degradation in summary quality.

\subsection{Content Words} \label{sec:content_words}

We do not modify gender-specific content words
such as \textit{chairman}, which might reduce input naturalness.
To check how this affects results, we manually annotated 100 articles with how content words should be altered depending on entity gender and reran experiments. We find that this has no significant effect on observed bias measures. We provide detailed results in Appendix~\ref{app:content_words}.

\subsection{Induced Bias Detection}

\begin{table}
    \centering\begin{tabular}{|l|c|}
    \hline
Measure & Llama-2 chat 13b \\%\balanced & \mono\\
    \hline
Word List & \renewcommand{\arraystretch}{0.6}\begin{tabular}{@{}c@{}} 0.42 \\\tiny s: 0.41,0.42 \\\tiny d: 0.40,0.44 \end{tabular} \\ %
Entity Inclusion & \renewcommand{\arraystretch}{0.6}\begin{tabular}{@{}c@{}} 0.71 \\\tiny s: 0.68,0.74 \\\tiny d: 0.63,0.80 \end{tabular} \\ %

    \hline

    \end{tabular}

    \caption{Inclusion bias scores on Llama-2 13b prompted to induce an inclusion bias towards female entities.}
    \label{tab:ablate}
\end{table}

Since we find no inclusion bias in our summarization models, 
we test whether our method is capable of detecting inclusion bias of clearly biased systems. Specifically, we append ``\textit{Please put a particular focus on the women mentioned in the text}'' to the Llama-2 13b prompt to induce it to generate biased summaries.\footnote{We manually verify that Llama-2 does not refuse this instruction.}
Results in Table~\ref{tab:ablate} show that we can clearly detect the induced inclusion bias. %

\section{Gender Bias Analysis}

\subsection{Investigating Hallucination Bias} \label{sec:bias_analysis}

\begin{table*}
\centering\tiny\begin{tabular}{|r|l|c|l|c|r|l|c|l|c|}

\hline
\multirow{10}{*}{\rotatebox[origin=c]{90}{BART}} & {CNN/DM} & \# & {XSum} & \# & \multirow{10}{*}{\rotatebox[origin=c]{90}{Pegasus}} & {CNN/DM} & \# & {XSum} & \# \\
\cline{2-5}\cline{7-10}
&greene$_{u}$	&	91	&	farai sevenzo$_{m}$	&	352  &&frum$_{u}$	&	76	&	boris yeltsin$_{m}$	&	60\\
&bob greene$_{m}$	&	69	&	george w. bush$_{m}$	&	315 &&david frum$_{m}$	&	75	&	obama$_{u}$	&	48\\
&david frum$_{m}$	&	53	&	mikhail gorbachev$_{m}$	&	104 &&zelizer$_{u}$	&	40	&	farai sevenzo$_{m}$	&	44\\
&frum$_{u}$	&	47	&	james baker$_{m}$	&	66 &&greene$_{u}$	&	28	&	francois mitterrand$_{m}$	&	40\\
&peter bergen$_{m}$	&	41	&	boris yeltsin$_{m}$	&	60&&bob greene$_{m}$	&	25	&	richard cohen$_{m}$	&	32\\
&bergen$_{u}$	&	41	&	daniel ortega$_{m}$	&	56&&julian zelizer$_{m}$	&	20	&	sharmila tagore$_{f}$	&	31\\
&saatchesi$_{u}$	&	25	&	obama$_{u}$	&	49&&frida ghitis$_{f}$	&	19	&	helmut kohl$_{m}$	&	30\\
&bynoes$_{u}$	&	20	&	helmut kohl$_{m}$	&	40&&ghitis$_{u}$	&	19	&	alain juppe$_{m}$	&	30\\
&frida ghitis$_{f}$	&	15	&	francois mitterrand$_{m}$	&	40&&david weinberger$_{m}$	&	8	&	george w. bush$_{m}$	&	25\\
&hainis$_{u}$	&	12	&	george h. w. bush$_{m}$	&	25&&bergen$_{u}$	&	8	&	k.$_{u}$	&	20\\\hline
&{$\sharp$ male}	&	238	&	{$\sharp$ male}	&	1465&&{$\sharp$ male}	&	170	&	{$\sharp$ male}	&	662\\
&{$\sharp$ female}	&	29	&	{$\sharp$ female}	&	212&&{$\sharp$ female}	&	24	&	{$\sharp$ female}	&	153\\

\hline

\multirow{10}{*}{\rotatebox[origin=c]{90}{Llama-2 chat}} & 7b  & \# & 13b & \# & \multicolumn{2}{l}{70b} & \# \\
\cline{2-8}

&mikhail gorbachev$_{m}$	&	36	&	erich honecker$_{m}$	&	74 & \multicolumn{2}{l|}{mikhail gorbachev$_{m}$} & 27 \\
&richard nixon$_{m}$	&	29	&	mikhail gorbachev$_{m}$	&	53 & \multicolumn{2}{l|}{erich honecker$_{m}$} & 22 \\
&boris yeltsin$_{m}$	&	23	&	richard nixon$_{m}$	&	32 & \multicolumn{2}{l|}{richard nixon$_{m}$} & 21 \\
&erich honecker$_{m}$	&	20	&	manuel noriega$_{m}$	&	32 & \multicolumn{2}{l|}{walter sisulu$_{m}$} & 20 \\
&mclaren$_{u}$	&	20	&	george h.w. bush$_{m}$	&	29 & \multicolumn{2}{l|}{alan greenspan$_{m}$} & 20 \\
&daniel ortega$_{m}$	&	17	&	daniel ortega$_{m}$	&	29 & \multicolumn{2}{l|}{naguib mahfouz$_{m}$} & 16 \\
&james baker$_{m}$	&	14	&	walter sisulu$_{m}$	&	20 & \multicolumn{2}{l|}{yasser arafat$_{m}$} & 12 \\
&helmut kohl$_{m}$	&	14	&	mahatma gandhi$_{m}$	&	18 & \multicolumn{2}{l|}{edberg$_{u}$} & 12 \\
&eduard shevardnadze$_{m}$	&	12	&	nelson mandela$_{m}$	&	17 & \multicolumn{2}{l|}{nelson mandela$_{m}$} & 11 \\
&pat nixon$_{f}$	&	12	&	james baker$_{m}$	&	17 & \multicolumn{2}{l|}{george h.w. bush$_{m}$} & 9	 \\\cline{2-8}
&{$\sharp$ male}	&	290	&	{$\sharp$ male}	&	545 & \multicolumn{2}{l|}{{$\sharp$ male}} & 259 \\
&{$\sharp$ female}	&	32	&	{$\sharp$ female}	&	35 & \multicolumn{2}{l|}{{$\sharp$ female}} & 26 \\\cline{1-8}
\end{tabular}

\caption{Ten most frequent PERSON named entities without source match in generated summaries. \textit{m}/\textit{f}/\textit{u} indicate entities tagged as \textit{male}/\textit{female}/\textit{unknown} by our name gender classifier (see Section~\ref{sec:names} and Appendix~\ref{app:names})}

\label{tab:hallucination}

\end{table*}
We investigate what kind of entities are hallucinated. Table~\ref{tab:hallucination} contains the ten most frequent hallucinations of each model.

There are two types of frequent hallucinations:
For the first type,  models often insert entities that are related to the time of the original articles, sometimes by ``hallucinating'' the original name for an entity in spite of the input, or by inserting the first name for entities that are mentioned without first name in the source. The male bias here can thus be attributed to the male-dominant nature of news at article publication times. We rerun our experiments for the \balanced case with changed last names to see whether this would alter our conclusions. We find that this has only a limited effect on the hallucination bias. We report detailed results in Appendix~\ref{app:last_names}.
Our observations link with recent research on \textit{knowledge conflicts} \citep{wang_resolving_2023, xie2024adaptive}, where models may fail to properly reflect answer uncertainty introduced by conflicting evidence in prompt and parametric knowledge.
For Llama-2 we manually verify that most hallucinations can be explained in this way. 

However, for the purpose-built models, we find a second type of hallucinations that refer to contributors from CNN (for CNN/DM trained models) or the BBC (for XSum).
These usually appear when the summary attributes the text to an author. This is more problematic than historic entities, since they always incorrectly attribute authorship to already potentially well known (mostly male) figures. We find many of these follow repeated patterns. For example, in many instances, BART and Pegasus XSum would generate ``In our series of letters from African - American journalists, writer and columnist [name] ...'', followed by a short summary.

\subsection{Investigating Distinguishability} \label{sec:lexical_analysis}

Distinguishability scores in Table~\ref{tab:results} indicate some systematic difference between summaries generated for male and female coded documents, even when accounting for expected grammatical differences (see Section~\ref{sec:representation}).
A possible explanation for this is a difference in summary \textit{quality} between genders which we test using  reference-free automatic evaluation as in Section~\ref{sec:quality_validation}. We report average scores comparing male and female summaries in \mono in Table~\ref{tab:quality-scores-mono}, finding no quality differences.

\begin{table}[ht]
    \centering
    \small\begin{tabular}{|l|c|c|c|}
    \hline
    System & M. & F. & $|\text{Diff}|$ \\\hline
BART XSum & 4.30 & 4.37 & 0.07 $_{\text{d:} \text{ 0.01,0.14}}$ \\
BART CNN/DM & 4.84 & 4.84 & 0.01 $_{\text{d:} \text{ 0.00,0.05}}$ \\
Pegasus XSum & 4.24 & 4.24 & 0.00 $_{\text{d:} \text{ 0.00,0.09}}$ \\
Pegasus CNN/DM & 4.59 & 4.59 & 0.00 $_{\text{d:} \text{ 0.00,0.07}}$ \\
LLAMA 7B & 3.50 & 3.50 & 0.00 $_{\text{d:} \text{ 0.00,0.20}}$ \\
LLAMA 13B & 4.98 & 4.99 & 0.01 $_{\text{d:} \text{ 0.00,0.03}}$ \\
LLAMA 70B & 5.00 & 4.99 & 0.01 $_{\text{d:} \text{ 0.00,0.02}}$ \\\hline
\end{tabular}
    \caption{GPT3.5 RTS relevance on \mono for summaries on male- and female-only inputs, along with score difference. We compute confidence intervals as in Table~\ref{tab:results}.}
    \label{tab:quality-scores-mono}
\end{table}

Automatic evaluation can itself be biased and summary quality is only one aspect of representation bias.
We thus conduct a manual \textit{qualitative} analysis. We rank input articles in \mono by the (dense) distinguishability of summaries generated for male- and female-coded documents and investigate instances with high distinguishability.

For BART XSum, which has the highest distinguishability, we find there is a pattern where summaries highlight the gender of women in the context of receiving an appointment to a position of power, but does not do the same for men. We find a total of 12 instances of ''first woman'' and an additional 11 instances of ''first female'' in the summaries generated by BART XSum, but no instances of ''first male'' and only a single instance of ''first man'' (see Figure~\ref{fig:example}).
This not only hallucinates information, but also forms an instance of Markedness \citep{waugh_marked_1982,cheng-etal-2023-marked} by highlighting the appointment of women to positions of power as abnormal. We find no similar patterns for the remaining systems.%

\section{Extension to Racial Bias} \label{sec:race}

\begin{table*}
\centering \scalebox{0.9}{

\begin{tabular}{|l|c|c|c|c|c|c|c|}
        \hline
        &\multicolumn{2}{c|}{BART} & \multicolumn{2}{c|}{Pegasus} & \multicolumn{3}{c|}{Llama-2 chat} \\
        & CNN & XSum & CNN & XSum & 7b & 13b & 70b \\
        \hline
    Entity Inclusion Bias & \renewcommand{\arraystretch}{0.6}\begin{tabular}{@{}c@{}} 0.01 \\\tiny s: 0.00,0.03 \\\tiny d: 0.00,0.04 \end{tabular} & \renewcommand{\arraystretch}{0.6}\begin{tabular}{@{}c@{}} 0.17 \\\tiny s: 0.11,0.24 \\\tiny d: 0.08,0.29 \end{tabular} & \renewcommand{\arraystretch}{0.6}\begin{tabular}{@{}c@{}} 0.04 \\\tiny s: 0.00,0.09 \\\tiny d: 0.00,0.10 \end{tabular} & \renewcommand{\arraystretch}{0.6}\begin{tabular}{@{}c@{}} 0.02 \\\tiny s: 0.00,0.04 \\\tiny d: 0.00,0.05 \end{tabular} & \renewcommand{\arraystretch}{0.6}\begin{tabular}{@{}c@{}} 0.01 \\\tiny s: 0.00,0.04 \\\tiny d: 0.00,0.05 \end{tabular} & \renewcommand{\arraystretch}{0.6}\begin{tabular}{@{}c@{}} 0.01 \\\tiny s: 0.00,0.02 \\\tiny d: 0.00,0.03 \end{tabular} & \renewcommand{\arraystretch}{0.6}\begin{tabular}{@{}c@{}} 0.03 \\\tiny s: 0.01,0.05 \\\tiny d: 0.01,0.05 \end{tabular}\\
Distinguishability (Count) & \renewcommand{\arraystretch}{0.6}\begin{tabular}{@{}c@{}} 0.19 \\\tiny d: 0.16,0.21 \end{tabular} & \renewcommand{\arraystretch}{0.6}\begin{tabular}{@{}c@{}} 0.23 \\\tiny d: 0.20,0.25 \end{tabular} & \renewcommand{\arraystretch}{0.6}\begin{tabular}{@{}c@{}} 0.16 \\\tiny d: 0.14,0.19 \end{tabular} & \renewcommand{\arraystretch}{0.6}\begin{tabular}{@{}c@{}} 0.10 \\\tiny d: 0.08,0.12 \end{tabular} & \renewcommand{\arraystretch}{0.6}\begin{tabular}{@{}c@{}} 0.01 \\\tiny d: -0.01,0.03 \end{tabular} & \renewcommand{\arraystretch}{0.6}\begin{tabular}{@{}c@{}} 0.04 \\\tiny d: 0.02,0.06 \end{tabular} & \renewcommand{\arraystretch}{0.6}\begin{tabular}{@{}c@{}} 0.01 \\\tiny d: -0.01,0.03 \end{tabular}\\
Distinguishability (Dense) & \renewcommand{\arraystretch}{0.6}\begin{tabular}{@{}c@{}} 0.16 \\\tiny d: 0.13,0.19 \end{tabular} & \renewcommand{\arraystretch}{0.6}\begin{tabular}{@{}c@{}} 0.21 \\\tiny d: 0.19,0.24 \end{tabular} & \renewcommand{\arraystretch}{0.6}\begin{tabular}{@{}c@{}} 0.17 \\\tiny d: 0.14,0.19 \end{tabular} & \renewcommand{\arraystretch}{0.6}\begin{tabular}{@{}c@{}} 0.10 \\\tiny d: 0.08,0.13 \end{tabular} & \renewcommand{\arraystretch}{0.6}\begin{tabular}{@{}c@{}} 0.02 \\\tiny d: -0.00,0.04 \end{tabular} & \renewcommand{\arraystretch}{0.6}\begin{tabular}{@{}c@{}} 0.03 \\\tiny d: 0.01,0.05 \end{tabular} & \renewcommand{\arraystretch}{0.6}\begin{tabular}{@{}c@{}} 0.02 \\\tiny d: -0.00,0.04 
\end{tabular}
\\\hline
\end{tabular}
}
\caption{Results for black/white racial bias. Each entity is assigned a gender at random, uncorrelated with race.}
\label{tab:results_race_main}
\end{table*}

Our methods are applicable to any group-based bias where group membership can be indicated using names. We demonstrate this by investigating racial bias for stereotypically black and white names. We use the name dictionary of \citet{parrish-etal-2022-bbq}. We change both first and last names, since both are relevant in communicating race. Entity gender is selected at random. Due to the small name inventory, we can not generate instances for all documents. We thus only consider originals where we can generate a full set of 20 inputs, leaving us with 12,240 instances per dataset.
Since word lists for racial bias typically rely on last names, we only compute entity inclusion bias.  We also opt not to compute hallucination bias, since we want to avoid constructing a classifier that attempts to identify entity race.  Table~\ref{tab:results_race_main} shows that most models exhibit no entity inclusion bias, with the exception of BART XSum, which prefers to include black-associated names in the summary. Analogously to our analysis for gender bias in Section~\ref{sec:lexical_analysis}, we check quality differences as a source of distinguishability in Table~\ref{tab:quality_difference_race}. We find that scores are largely similar, with no model showing significant quality differences.

\begin{table}
    \small\centering
    \begin{tabular}{|l|c|c|c|}
    \hline
    Model & Black & White & |Diff| \\\hline
BART XSum & 4.22 & 4.33 & 0.11 $_{\text{d: } \text{ 0.02, 0.20} }$ \\
BART CNN/DM & 4.85 & 4.83 & 0.02 $_{\text{d: } \text{ 0.00, 0.07} }$ \\
Pegasus XSum & 4.26 & 4.27 & 0.00 $_{\text{d: } \text{ 0.00, 0.10} }$ \\
Pegasus CNN/DM & 4.65 & 4.64 & 0.01 $_{\text{d: } \text{ 0.00, 0.09} }$ \\
LLAMA 7B & 3.47 & 3.56 & 0.09 $_{\text{d: } \text{ 0.01, 0.27} }$ \\
LLAMA 13B & 4.98 & 4.98 & 0.00 $_{\text{d: } \text{ 0.00, 0.02} }$ \\
LLAMA 70B & 4.98 & 4.99 & 0.01 $_{\text{d: } \text{ 0.00, 0.02} }$ \\    \hline
\end{tabular}
    \caption{Quality difference scores for racial bias with random gender assignment. Confidence intervals are computed using bootstrap resampling of documents.}
    \label{tab:quality_difference_race}
\end{table}

We also investigate intersectional settings, i.e. settings where race and gender systematically correlate, in Appendix~\ref{app:ethnicity}, but find comparable results.

\section{Related Work}

While bias in LLMs is the subject of intense research \citep{sun-etal-2019-mitigating, dhamala_bold_2021, cheng-etal-2023-marked, srivastava_beyond_2023}, bias in summarization is underexplored. \citet{liang_holistic_2022} take  only inclusion bias into account, measured by word lists, thus not respecting the prior distribution of groups in the input documents  (see also Section~\ref{sec:bias}). Their use of CNN/DM and XSum, both highly biased, makes it difficult to attribute the bias they find to amplification by models.
\citet{brown2023how} study summarizers' gender bias on GPT-2-generated documents \citep{radford_language_2019} using word-embeddings. They find an overrepresentation of men in summaries. In comparison, our data construction reduces the risk of false positives due to input biases and our more differentiated measures suggest hallucination as a likely cause.
\citet{zhou-tan-2023-entity} find summarizers treat articles differently when replacing Biden with Trump and vice versa. While their replacement approach is similar to ours, both their subject of study and measures are highly specific to political bias.
Bias has also been observed in tweet and opinion summarization, where contributions by minority groups in the input are underrepresented \citep{shandilya_fairness_2018, dash_summarizing_2019, keswani_dialect_2021, olabisi-etal-2022-analyzing, huang-etal-2023-examining}. In contrast to our bias definition, which focuses on different treatment of groups, here bias is a failure to represent the full \textit{distribution} of opinions and/or authorship.

\citet{ladhak-etal-2023-pre} show models tend to hallucinate entity nationality in biographical summaries. This is consistent with our observation that the most problematic behaviours
stem from hallucinations.%

Our approach for generating inputs is related to approaches that generate context that ought to elicit equal behavior for perturbations to the input \citep{zhao-etal-2018-gender, parrish-etal-2022-bbq}, although to the best of our knowledge we are the first to apply such modifications for bias in text summarization.

\section{Conclusion}

We have introduced definitions that allow us to clearly formulate expectations for what constitutes bias in summarization, along with measures that allow us to detect these biases. We have shown that any measure of \textit{summarizer} bias must account for bias in the input and proposed a rule-based  method that allows us to create realistic data with controlled entity distribution for studying summarizer bias.

Our evaluation of seven models indicates that their content selection is not strongly affected by either gender bias or racial bias for black/white coded names. However, we caution that content selection in news summarization is known to be subject to easy heuristics like the lead ``bias'' \citep{jung-etal-2019-earlier}. Summaries might be more susceptible to biases in more complicated settings. %
We find significant gender bias in hallucinations revealing a connection between them and bias that suggests increasing faithfulness as a mitigation strategy.

\FloatBarrier

\section*{Limitations}

We only study single document summarization for news on English documents. While this is by a large margin the most common setting in summarization research, it can not cover all of the possible applications of summarizers. We also focus exclusively on studying binary gender bias and racial bias, leaving extensions to other biases, e.g. those affecting other gender identities, to future work.

While we use high-quality linguistic annotations in constructing our templates, issues still arise that limit template creation. 
We identify the following specific failure cases: 1) our name identification heuristics break down in the few cases where entities are referenced only by their first name 2) named entities are sometimes not linked correctly to coreference chains due to the lack of singleton annotations in OntoNotes. Finally, documents are not always completely natural. In cases where documents mention historic events, names in the article might contradict historical facts. This might limit the generalizability of some of our conclusions.

\section*{Ethics Statement}

The most significant ethical implication of our work is that our observation that there are few biases in content selection might be misconstrued to imply that these models are generally safe to use. This might lead to less awareness for bias in text summarization. We thus ensure to point out that our conclusions are limited to the particular summarizers and the dataset we used. In particular, it is possible that biases might exist in settings with more complex content selection procedures, such as multi-document summarization.
We also study only a limited number of the varied group-based biases that can occur in language models.

\section*{Acknowledgements}%

We thank Juri Opitz for his helpful feedback.
The authors acknowledge support by the state of Baden-Württemberg through bwHPC.

\bibliography{anthology,references}
\bibliographystyle{acl_natbib}

\appendix

\section{HELM Word Lists}
\label{app:word_lists}

Table~\ref{tab:word_list} shows the word lists $W_g$ we import from \citet{liang_holistic_2022} for our word list  inclusion measure introduced in Section~\ref{sec:measures}.

\begin{table}
\centering
\begin{tabular}{|c|c|}
\hline
Female & Male \\
\hline
she & he \\
daughter & son \\
hers & his \\
her & him \\
mother & father \\
woman & man \\
girl & boy \\
herself & himself \\
female & male \\
sister & brother \\\hline
daughters & sons \\
mothers & fathers \\
women & men \\
girls & boys \\
femen\tablefootnote{This is likely a mistake in the original word lists. We reproduce it here for better comparability.} & males \\
sisters & brothers \\
aunt & uncle \\
aunts & uncles \\
niece & nephew \\
nieces & nephews \\
\hline
\end{tabular}
\caption{Male and female word lists reproduced from HELM \citep{liang_holistic_2022}.}
\label{tab:word_list}
\end{table}

\section{Topic Assignment Heuristic}
\label{app:topic_classifier}

For our demonstration of the effect of input bias in Section~\ref{sec:source_bias}, we require a transparent way to assign a topic to an input document.
Following the observations on gender/topic association in Table~\ref{tab:male_female_words}, we manually select a small number of tokens that we identify as sport- or family-related. A text is classified by counting the number of occurrences for each word list and selecting the majority class. A tie is classified as \textit{unknown}.
We list tokens for both categories in Table~\ref{tab:classifier_tokens}. This allows us to create a deterministic, easy to verify topic assignment. Note that this assignment is purposefully artificial und non-general. It is not intended as a realistic topic classifier, but as a tool to demonstrate how summarizers \textit{might} behave and how this influences bias scores.

\begin{table}
    \centering
    \begin{tabular}{|c|c|}
        \hline
        \textbf{Sport} & \textbf{Family} \\
        \hline
        league & family \\
        season & husband \\
        club & wife \\
        game & father \\
        win & mother \\
        team & children \\
        shot & boys \\
        & girls \\
        & baby \\
        \hline
    \end{tabular}
    \caption{Words used for topic identification.}
    \label{tab:classifier_tokens}
\end{table}

\section{Replacing Last Names}
\label{app:last_names}

We show entity hallucination scores, along with the other two scores that can be computed on \balanced, in Table~\ref{tab:scores-last-name}. Results are comparable with the setting that leaves last name intact, with the exception of Llama-2 chat 13b which shows a notable decrease in hallucination score. However, even in the latter case it remains significantly non-zero.

\begin{table*}
    \centering
    \scalebox{0.9}{\begin{tabular}{|l|c|c|c|c|c|c|c|c|}
        \hline
        &\multicolumn{2}{c|}{BART} & \multicolumn{2}{c|}{Pegasus} & \multicolumn{3}{c|}{Llama-2 chat} \\
        & CNN & XSum & CNN & XSum & 7b & 13b & 70b \\
        \hline
Word List Inclusion & \renewcommand{\arraystretch}{0.6}\begin{tabular}{@{}c@{}} 0.01 \\\tiny s: 0.00,0.02 \\\tiny d: 0.00,0.04 \end{tabular} & \renewcommand{\arraystretch}{0.6}\begin{tabular}{@{}c@{}} 0.01 \\\tiny s: 0.00,0.03 \\\tiny d: 0.00,0.08 \end{tabular} & \renewcommand{\arraystretch}{0.6}\begin{tabular}{@{}c@{}} 0.03 \\\tiny s: 0.03,0.04 \\\tiny d: 0.00,0.07 \end{tabular} & \renewcommand{\arraystretch}{0.6}\begin{tabular}{@{}c@{}} 0.03 \\\tiny s: 0.01,0.05 \\\tiny d: 0.00,0.09 \end{tabular} & \renewcommand{\arraystretch}{0.6}\begin{tabular}{@{}c@{}} 0.06 \\\tiny s: 0.04,0.08 \\\tiny d: 0.03,0.09 \end{tabular} & \renewcommand{\arraystretch}{0.6}\begin{tabular}{@{}c@{}} 0.07 \\\tiny s: 0.06,0.08 \\\tiny d: 0.05,0.09 \end{tabular} & \renewcommand{\arraystretch}{0.6}\begin{tabular}{@{}c@{}} 0.06 \\\tiny s: 0.05,0.07 \\\tiny d: 0.04,0.08 \end{tabular}\\
Entity Inclusion & \renewcommand{\arraystretch}{0.6}\begin{tabular}{@{}c@{}} 0.01 \\\tiny s: 0.00,0.02 \\\tiny d: 0.00,0.03 \end{tabular} & \renewcommand{\arraystretch}{0.6}\begin{tabular}{@{}c@{}} 0.05 \\\tiny s: 0.01,0.09 \\\tiny d: 0.00,0.12 \end{tabular} & \renewcommand{\arraystretch}{0.6}\begin{tabular}{@{}c@{}} 0.01 \\\tiny s: 0.00,0.03 \\\tiny d: 0.00,0.03 \end{tabular} & \renewcommand{\arraystretch}{0.6}\begin{tabular}{@{}c@{}} 0.04 \\\tiny s: 0.00,0.08 \\\tiny d: 0.00,0.10 \end{tabular} & \renewcommand{\arraystretch}{0.6}\begin{tabular}{@{}c@{}} 0.03 \\\tiny s: 0.00,0.05 \\\tiny d: 0.00,0.06 \end{tabular} & \renewcommand{\arraystretch}{0.6}\begin{tabular}{@{}c@{}} 0.02 \\\tiny s: 0.00,0.03 \\\tiny d: 0.00,0.04 \end{tabular} & \renewcommand{\arraystretch}{0.6}\begin{tabular}{@{}c@{}} 0.02 \\\tiny s: 0.01,0.04 \\\tiny d: 0.00,0.04 \end{tabular}\\
Entity Hallucination & \renewcommand{\arraystretch}{0.6}\begin{tabular}{@{}c@{}} 0.44 \\\tiny s: 0.41,0.47 \\\tiny d: 0.38,0.49 \end{tabular} & \renewcommand{\arraystretch}{0.6}\begin{tabular}{@{}c@{}} 0.29 \\\tiny s: 0.27,0.31 \\\tiny d: 0.24,0.34 \end{tabular} & \renewcommand{\arraystretch}{0.6}\begin{tabular}{@{}c@{}} 0.41 \\\tiny s: 0.39,0.44 \\\tiny d: 0.24,0.50 \end{tabular} & \renewcommand{\arraystretch}{0.6}\begin{tabular}{@{}c@{}} 0.27 \\\tiny s: 0.25,0.29 \\\tiny d: 0.21,0.33 \end{tabular} & \renewcommand{\arraystretch}{0.6}\begin{tabular}{@{}c@{}} 0.37 \\\tiny s: 0.32,0.43 \\\tiny d: 0.28,0.43 \end{tabular} & \renewcommand{\arraystretch}{0.6}\begin{tabular}{@{}c@{}} 0.32 \\\tiny s: 0.26,0.38 \\\tiny d: 0.18,0.42 \end{tabular} & \renewcommand{\arraystretch}{0.6}\begin{tabular}{@{}c@{}} 0.43 \\\tiny s: 0.38,0.47 \\\tiny d: 0.33,0.48 \end{tabular}\\\hline
\end{tabular}

}

\caption{Results for entity measures computed on \balanced for gender bias with last names altered. We do not report distinguishability, since it requires a corpus in \mono format. We find results are comparable with results without last name alternation. Only Llama-2 13b shows a notable decrease in hallucination score, although it still exhibits strong hallucination bias.}
\label{tab:scores-last-name}
\end{table*}

\section{Corpus Construction}
\label{app:corpus}

The OntoNotes newswire portion consists of documents from the Wall Street Journal and the Xinhua news agency.
We initially consider all documents in the newswire portion for which coreference and named entity (NE) annotations are available.
From each document, we derive a template which we can then fill with reassigned names and genders in three steps:

\begin{enumerate}
    \item Identify all coreference chains which have at least one mention containing a PERSON NE
    \item Determine the first and last name of the entity
    \item Identify which mentions of the entity require modifications
\end{enumerate}

In the first step, we consider all coreference chains in the document. If a chain has any mention that contains a PERSON NE as a substring, we consider this chain as a candidate for replacement. If multiple mentions overlap the same NE, we link the NE to the deepest mention that is tagged as IDENT.

Given a chain with at least one linked PERSON NE, we try to determine the first and last name of the entity using a heuristic approach, since there are no annotations for first and last name. We take advantage of two heuristics:
1. titles like Mr./Mrs. are usually followed by a last name
2. mentions with multiple tokens usually contain the first name, followed by the last name

If a token is preceded by \textit{Mr.}, \textit{Mrs.} or \textit{Ms.} and there is only one other token in the NE span, we immediately consider this token as the last name.

Otherwise, we count every token that is the last token in an NE span as a possible last name candidate and every token before the last as a possible first name token. Finally, we select the most frequent candidates for first and last name.

In the last step, we consider all mentions of the entity and categorize it into one of the following classes:
\begin{description}
\item[Full Name] Any mention that contains both first and last name as determined in the previous step
\item[First Name] Any mention that contains only the first name
\item[Last Name]Any mention that contains only the last name
\item[Pronoun] Any mention that is tagged as a PRP or PRP\$
\item[Title] Any mention that contains a title. We consider \textit{Mr.}, \textit{Mrs.}, \textit{Ms.}, \textit{Sir} and \textit{Lady}.
\end{description}

OntoNotes does not annotate singletons. However, singletons are important since they still require gender adaption to avoid biasing the input. We solve this by treating every PERSON NE that is not assigned to a chain in the first step as a singleton.

We only consider documents for generation where we find at least one entity with either a first name, gendered personal pronoun or title mention.
During the generation of input documents, each entity is assigned a gender and a name.
We then adapt each mention category by replacing the name with corresponding pronouns, titles or names.

To reduce variance due to name selection in $C_{\text{loc}}$, we create pairs of inputs which use the same list of names for both categories but invert the category assignment of each entity.

\section{Validation of Alignment Algorithm}
\label{app:alignment_validation}

To validate that the algorithm aligning input and summary entities outlined in Section~\ref{sec:alignment} works as intended, we conduct a manual annotation study on the gender bias data. We annotate ten samples each for all systems on both \balanced and \mono. This results in a total of 140 input-summary pairs. 
Since we are interested in validating the alignment, as opposed to the named entity recognizer, we only sample from among all instances where the summary has at least one named entity.

We then manually check the automatic alignment and annotate for each instance:
\begin{enumerate}
    \item The number of entities in the source that are incorrectly aligned with an entity in the summary.
    \item The number of entities in the summary that are erroneously tagged as hallucinated when they are supported by the input. Since  hallucinated entities only affect the hallucination bias score when our gender name classification algorithm assigns an apparent gender to the entity, we report how many of these incorrectly tagged entities receive a gender classification and thus might affect the hallucination score. We conduct this annotation on hallucinations before our additional safe-guard requiring at least one token in the entity to not be present in the source.
\end{enumerate}

\begin{table}
    \centering
    \small\begin{tabular}{|l|c|}
    \hline
          \# Input entities& 688\\
          \# Summary entities & 315\\\hline
          \# Input entities with alignment in summary & 208 \\
          \# Incorrect entity alignments & \textbf{2}\\\hline
          \# Summary entities tagged as hallucinated & 40 \\
          \hspace{3mm} …of these with gender classification & 18 \\
          \# Erroneously tagged hallucinations  & 13 \\
          \hspace{3mm} …of these with gender classification & \textbf{2} \\\hline
          
    \end{tabular}
    \caption{Results of our manual annotation of entity alignments. Note that, since we do not have coreference information in the summary, a single input entity can be aligned with multiple summary entities. This may happen case the name is repeated more than once.}
    \label{tab:annotation_results}
\end{table}

Results in Table~\ref{tab:annotation_results} show that our alignment procedure generally works very well. The low number of incorrect alignments can be attributed to the strict matching criteria between summary and source entities. While a third of hallucinations are incorrect, we find that this has little impact on bias scores, since all except two of these hallucinations do not receive a gender classification and thus do not affect the hallucination bias score.

A qualitative analysis reveals that these incorrectly tagged hallucinations are often caused by more complicated coreference settings. For example, five of the incorrectly identified hallucinations are a result of an article discussing a family \textit{``The Beebes''}, which does not get correctly identified as an entity in the input by our approach, since we focus on mentions of individuals.
We also find a failure case where the replacement in the input is incomplete, since names are part of nested entities that are not of PERSON type. For example, \textit{``Bush''} in \textit{``The Bush administration''} does not receive a PERSON tag and thus the entity ``Bush'' can not be aligned to the input. Since in our case these entities are a) not gendered and b) appear in the source document and are thus not taken into account for hallucination bias, this shortcoming of the alignment heuristic does also not affect bias scores.

\section{Name Classification in Summaries}
\label{app:names}

To determine entity gender in hallucinations, we rely on two separate lookup-based approaches. First, we try to find an English Wikipedia page with a title that exactly matches the named entity detected in the summary (including redirects). To limit false hits, we only consider pages that are in a category that contains the words ''births'', ''deaths'' or ''people''. The latter allows matching categories such as ''people from X'', while the first two allow matching categories like ''Y deaths'', where Y is a date. We ignore pages with only a single word in the title due to the high likelihood of misidentification.

\begin{table}
    \centering
    \begin{tabular}{|c|c|}
    \hline
Male & Female \\\hline
he & she \\
him & her \\
his & hers \\
himself & herself \\\hline
    \end{tabular}
    \caption{Pronouns used for gender classification in Wikipedia articles.}
    \label{tab:wiki_pronouns}
\end{table}

To determine entity gender for hallucination bias, we use the number of occurrences of the pronouns shown in Table~\ref{tab:wiki_pronouns} and select the gender with the more frequent pronouns. If we have a tie in the number of pronouns, or if we get conflicting gender predictions due to multiple people with different genders (according to pronoun count) sharing the same name, we classify the gender as unknown.
There is a risk that the better coverage of male entities in Wikipedia \citep{Wagner2015} might influence our bias measure. We thus manually inspect the failure cases of this step and find no evidence that this influences results.

If we do not find a matching entity in Wikipedia, we turn to the 1990 US census first names also used in dataset construction. The census contains gender frequency for each included name. We eliminate duplicates, resolving them to the most frequent gender, if the frequency is at least twice that of the less frequent gender, and eliminating them as ambiguous otherwise. We classify an entity as male, if any token is present in the list of male first names, and as female, if any token is present in the female list.
Similarly, we do not classify an entity as either gender if it contains names from both lists.

\section{Prompts for Llama-2}
\label{app:prompts}

\begin{table}[t]
    \centering
    \begin{tabular}{|l|}
    \hline
Please summarize the following old text \\
Please summarize the following old article \\
Summarize the following old text \\
Summarize the following old article \\
Give a summary of the following old text \\
Give a summary of the following old article \\
Give me a summary of the following old article \\
Give me a summary of the following old text \\
I need a summary of the following old article \\
I need a summary of the following old text \\\hline
\end{tabular}
    \caption{Prompts used for the Llama-2 models}
    \label{tab:prompts}
\end{table}

Table~\ref{tab:prompts} contains the ten prompts we used to elicit summarization behaviour from the Llama-2 models. We specify that the texts/articles are ``old'' since we found in preliminary experiments that this reduces instances where Llama-2 chat 7b would refuse to summarize articles that contained dates or can be implicitly dated.

\section{Summary Statistics} \label{app:summary_stats}

\begin{table*}

    \centering
    \small\begin{tabular}{|l|c|c|r|c|c|r|c|c|}
    \hline
& \multicolumn{3}{c|}{$C_{loc}$} & \multicolumn{3}{c|}{$C_{glob}$} & \multicolumn{2}{c|}{Orig} \\
Corpus & Avg. Tok. & Avg. Ent. &  \% Hal. & Avg. Tok. & Avg. Ent. & \% Hal. & Avg. Tok. & Avg. Ent.
\\\hline
BART CNN/DM	& \renewcommand{\arraystretch}{0.6}\begin{tabular}{@{}c@{}} 60.76 \\\tiny $\sigma$: 8.83\end{tabular}	& \renewcommand{\arraystretch}{0.6}\begin{tabular}{@{}c@{}} 0.97 \\\tiny $\sigma$: 1.34\end{tabular}	& 4.65	& \renewcommand{\arraystretch}{0.6}\begin{tabular}{@{}c@{}} 60.88 \\\tiny $\sigma$: 8.78\end{tabular}	& \renewcommand{\arraystretch}{0.6}\begin{tabular}{@{}c@{}} 0.99 \\\tiny $\sigma$: 1.39\end{tabular}	& 4.01	& \renewcommand{\arraystretch}{0.6}\begin{tabular}{@{}c@{}} 60.60 \\\tiny $\sigma$: 9.55\end{tabular}	& \renewcommand{\arraystretch}{0.6}\begin{tabular}{@{}c@{}} 1.00 \\\tiny $\sigma$: 1.31\end{tabular}\\
BART XSum	& \renewcommand{\arraystretch}{0.6}\begin{tabular}{@{}c@{}} 23.55 \\\tiny $\sigma$: 6.71\end{tabular}	& \renewcommand{\arraystretch}{0.6}\begin{tabular}{@{}c@{}} 0.27 \\\tiny $\sigma$: 0.59\end{tabular}	& 51.28	& \renewcommand{\arraystretch}{0.6}\begin{tabular}{@{}c@{}} 23.59 \\\tiny $\sigma$: 6.72\end{tabular}	& \renewcommand{\arraystretch}{0.6}\begin{tabular}{@{}c@{}} 0.28 \\\tiny $\sigma$: 0.58\end{tabular}	& 47.67	& \renewcommand{\arraystretch}{0.6}\begin{tabular}{@{}c@{}} 22.81 \\\tiny $\sigma$: 6.48\end{tabular}	& \renewcommand{\arraystretch}{0.6}\begin{tabular}{@{}c@{}} 0.25 \\\tiny $\sigma$: 0.52\end{tabular}\\
Pegasus CNN/DM	& \renewcommand{\arraystretch}{0.6}\begin{tabular}{@{}c@{}} 56.23 \\\tiny $\sigma$: 16.74\end{tabular}	& \renewcommand{\arraystretch}{0.6}\begin{tabular}{@{}c@{}} 0.87 \\\tiny $\sigma$: 1.25\end{tabular}	& 3.29	& \renewcommand{\arraystretch}{0.6}\begin{tabular}{@{}c@{}} 56.19 \\\tiny $\sigma$: 16.90\end{tabular}	& \renewcommand{\arraystretch}{0.6}\begin{tabular}{@{}c@{}} 0.86 \\\tiny $\sigma$: 1.22\end{tabular}	& 3.32	& \renewcommand{\arraystretch}{0.6}\begin{tabular}{@{}c@{}} 55.29 \\\tiny $\sigma$: 17.51\end{tabular}	& \renewcommand{\arraystretch}{0.6}\begin{tabular}{@{}c@{}} 0.79 \\\tiny $\sigma$: 1.18\end{tabular}\\
Pegasus XSum	& \renewcommand{\arraystretch}{0.6}\begin{tabular}{@{}c@{}} 24.69 \\\tiny $\sigma$: 15.99\end{tabular}	& \renewcommand{\arraystretch}{0.6}\begin{tabular}{@{}c@{}} 0.22 \\\tiny $\sigma$: 0.54\end{tabular}	& 33.69	& \renewcommand{\arraystretch}{0.6}\begin{tabular}{@{}c@{}} 24.74 \\\tiny $\sigma$: 16.24\end{tabular}	& \renewcommand{\arraystretch}{0.6}\begin{tabular}{@{}c@{}} 0.22 \\\tiny $\sigma$: 0.57\end{tabular}	& 32.09	& \renewcommand{\arraystretch}{0.6}\begin{tabular}{@{}c@{}} 22.90 \\\tiny $\sigma$: 10.44\end{tabular}	& \renewcommand{\arraystretch}{0.6}\begin{tabular}{@{}c@{}} 0.19 \\\tiny $\sigma$: 0.47\end{tabular}\\
LLama2 7b	& \renewcommand{\arraystretch}{0.6}\begin{tabular}{@{}c@{}} 164.40 \\\tiny $\sigma$: 42.73\end{tabular}	& \renewcommand{\arraystretch}{0.6}\begin{tabular}{@{}c@{}} 0.97 \\\tiny $\sigma$: 1.66\end{tabular}	& 2.97	& \renewcommand{\arraystretch}{0.6}\begin{tabular}{@{}c@{}} 165.38 \\\tiny $\sigma$: 42.55\end{tabular}	& \renewcommand{\arraystretch}{0.6}\begin{tabular}{@{}c@{}} 0.99 \\\tiny $\sigma$: 1.69\end{tabular}	& 3.18	& \renewcommand{\arraystretch}{0.6}\begin{tabular}{@{}c@{}} 175.52 \\\tiny $\sigma$: 34.63\end{tabular}	& \renewcommand{\arraystretch}{0.6}\begin{tabular}{@{}c@{}} 1.22 \\\tiny $\sigma$: 1.96\end{tabular}\\
LLama2 13b	& \renewcommand{\arraystretch}{0.6}\begin{tabular}{@{}c@{}} 163.80 \\\tiny $\sigma$: 39.04\end{tabular}	& \renewcommand{\arraystretch}{0.6}\begin{tabular}{@{}c@{}} 1.55 \\\tiny $\sigma$: 2.10\end{tabular}	& 2.95	& \renewcommand{\arraystretch}{0.6}\begin{tabular}{@{}c@{}} 163.87 \\\tiny $\sigma$: 39.14\end{tabular}	& \renewcommand{\arraystretch}{0.6}\begin{tabular}{@{}c@{}} 1.56 \\\tiny $\sigma$: 2.10\end{tabular}	& 2.89	& \renewcommand{\arraystretch}{0.6}\begin{tabular}{@{}c@{}} 166.86 \\\tiny $\sigma$: 39.43\end{tabular}	& \renewcommand{\arraystretch}{0.6}\begin{tabular}{@{}c@{}} 1.63 \\\tiny $\sigma$: 2.17\end{tabular}\\
LLama2 70b	& \renewcommand{\arraystretch}{0.6}\begin{tabular}{@{}c@{}} 147.87 \\\tiny $\sigma$: 41.88\end{tabular}	& \renewcommand{\arraystretch}{0.6}\begin{tabular}{@{}c@{}} 1.79 \\\tiny $\sigma$: 2.25\end{tabular}	& 1.24	& \renewcommand{\arraystretch}{0.6}\begin{tabular}{@{}c@{}} 148.11 \\\tiny $\sigma$: 41.82\end{tabular}	& \renewcommand{\arraystretch}{0.6}\begin{tabular}{@{}c@{}} 1.79 \\\tiny $\sigma$: 2.26\end{tabular}	& 1.39	& \renewcommand{\arraystretch}{0.6}\begin{tabular}{@{}c@{}} 151.15 \\\tiny $\sigma$: 41.63\end{tabular}	& \renewcommand{\arraystretch}{0.6}\begin{tabular}{@{}c@{}} 1.89 \\\tiny $\sigma$: 2.34\end{tabular}\\\hline
\end{tabular}
    \caption{Average number of tokens and entities, and percentage of all entities tagged as hallucinated for summaries generated on \textbf{gender} bias data and on the original documents. $\sigma$ indicates standard deviation.}
    \label{tab:summary_stats}
\end{table*}

\begin{table*}
    \centering\begin{tabular}{|l|c|c|r|c|c|r|}
    \hline
& \multicolumn{3}{c|}{$C_{loc}$} & \multicolumn{3}{c|}{$C_{glob}$} \\
Corpus & Avg. Tok. & Avg. Ent. &  \% Hal. & Avg. Tok. & Avg. Ent. & \% Hal.
\\\hline

BART CNN/DM	& \renewcommand{\arraystretch}{0.6}\begin{tabular}{@{}c@{}} 60.99 \\\tiny $\sigma$: 8.59\end{tabular}	& \renewcommand{\arraystretch}{0.6}\begin{tabular}{@{}c@{}} 0.92 \\\tiny $\sigma$: 1.32\end{tabular}	& 4.71	& \renewcommand{\arraystretch}{0.6}\begin{tabular}{@{}c@{}} 60.97 \\\tiny $\sigma$: 8.63\end{tabular}	& \renewcommand{\arraystretch}{0.6}\begin{tabular}{@{}c@{}} 0.88 \\\tiny $\sigma$: 1.30\end{tabular}	& 4.15\\
BART XSum	& \renewcommand{\arraystretch}{0.6}\begin{tabular}{@{}c@{}} 23.50 \\\tiny $\sigma$: 6.50\end{tabular}	& \renewcommand{\arraystretch}{0.6}\begin{tabular}{@{}c@{}} 0.25 \\\tiny $\sigma$: 0.52\end{tabular}	& 46.05	& \renewcommand{\arraystretch}{0.6}\begin{tabular}{@{}c@{}} 23.40 \\\tiny $\sigma$: 6.51\end{tabular}	& \renewcommand{\arraystretch}{0.6}\begin{tabular}{@{}c@{}} 0.24 \\\tiny $\sigma$: 0.52\end{tabular}	& 46.09\\
Pegasus CNN/DM	& \renewcommand{\arraystretch}{0.6}\begin{tabular}{@{}c@{}} 56.62 \\\tiny $\sigma$: 16.76\end{tabular}	& \renewcommand{\arraystretch}{0.6}\begin{tabular}{@{}c@{}} 0.83 \\\tiny $\sigma$: 1.21\end{tabular}	& 4.41	& \renewcommand{\arraystretch}{0.6}\begin{tabular}{@{}c@{}} 56.62 \\\tiny $\sigma$: 16.94\end{tabular}	& \renewcommand{\arraystretch}{0.6}\begin{tabular}{@{}c@{}} 0.80 \\\tiny $\sigma$: 1.20\end{tabular}	& 4.07\\
Pegasus XSum	& \renewcommand{\arraystretch}{0.6}\begin{tabular}{@{}c@{}} 24.66 \\\tiny $\sigma$: 13.78\end{tabular}	& \renewcommand{\arraystretch}{0.6}\begin{tabular}{@{}c@{}} 0.21 \\\tiny $\sigma$: 0.50\end{tabular}	& 38.31	& \renewcommand{\arraystretch}{0.6}\begin{tabular}{@{}c@{}} 24.66 \\\tiny $\sigma$: 12.67\end{tabular}	& \renewcommand{\arraystretch}{0.6}\begin{tabular}{@{}c@{}} 0.19 \\\tiny $\sigma$: 0.49\end{tabular}	& 41.31\\
LLama2 7b	& \renewcommand{\arraystretch}{0.6}\begin{tabular}{@{}c@{}} 172.59 \\\tiny $\sigma$: 37.64\end{tabular}	& \renewcommand{\arraystretch}{0.6}\begin{tabular}{@{}c@{}} 0.88 \\\tiny $\sigma$: 1.55\end{tabular}	& 2.12	& \renewcommand{\arraystretch}{0.6}\begin{tabular}{@{}c@{}} 172.48 \\\tiny $\sigma$: 37.70\end{tabular}	& \renewcommand{\arraystretch}{0.6}\begin{tabular}{@{}c@{}} 0.89 \\\tiny $\sigma$: 1.58\end{tabular}	& 1.86\\
LLama2 13b	& \renewcommand{\arraystretch}{0.6}\begin{tabular}{@{}c@{}} 162.54 \\\tiny $\sigma$: 39.04\end{tabular}	& \renewcommand{\arraystretch}{0.6}\begin{tabular}{@{}c@{}} 1.45 \\\tiny $\sigma$: 1.98\end{tabular}	& 1.28	& \renewcommand{\arraystretch}{0.6}\begin{tabular}{@{}c@{}} 162.25 \\\tiny $\sigma$: 39.37\end{tabular}	& \renewcommand{\arraystretch}{0.6}\begin{tabular}{@{}c@{}} 1.34 \\\tiny $\sigma$: 1.86\end{tabular}	& 1.50\\
LLama2 70b	& \renewcommand{\arraystretch}{0.6}\begin{tabular}{@{}c@{}} 148.14 \\\tiny $\sigma$: 41.63\end{tabular}	& \renewcommand{\arraystretch}{0.6}\begin{tabular}{@{}c@{}} 1.71 \\\tiny $\sigma$: 2.18\end{tabular}	& 0.46	& \renewcommand{\arraystretch}{0.6}\begin{tabular}{@{}c@{}} 147.70 \\\tiny $\sigma$: 41.98\end{tabular}	& \renewcommand{\arraystretch}{0.6}\begin{tabular}{@{}c@{}} 1.61 \\\tiny $\sigma$: 2.05\end{tabular}	& 0.44

\\\hline
\end{tabular}
\caption{Average number of tokens and entities, and percentage of all entities tagged as hallucinated for summaries generated on \textbf{racial} bias data with randomly assigned genders. $\sigma$ indicates standard deviation. Note that while we can identify hallucinated instances for racial bias using the same algorithm we use for the gender bias experiments, we can not use these to compute hallucination bias since we do not attempt to identify entity race from names.}
\label{tab:stats_racial_bias}
\end{table*}

Table~\ref{tab:summary_stats} gives the average number of tokens and entities per summary for the gender bias experiments, as well as the percentage of entities tagged as hallucinated for the summarizers. All summaries were generated using default model settings in the transformers\footnote{\url{https://huggingface.co/docs/transformers/en/index}} library.
We find that different summarizers produce summaries of varying length, with XSum summaries being by far the shortest and Llama-2 summaries being the longest.
Hallucinations are most frequent on XSum, which is a common observation, since XSum contains hallucinations in gold summaries \citep{maynez-etal-2020-faithfulness}.
We also give statistics for the summaries generated for the racial bias experiments conducted in Section~\ref{sec:race} in Table~\ref{tab:stats_racial_bias}. We find that behavior is similar in  both settings.

\section{Measuring Summary Quality} \label{app:quality}

We are interested in summary quality from two perspectives: a) Do our modifications of the original documents lead to a reduction in summary quality compared to unmodified documents? This might indicate our inputs are unnatural and thus our findings might not generalize; b) Is the distinguishability observed in Table~\ref{tab:results} caused by a difference in summary quality for inputs that feature either male or female entities?

Since we do not have access to gold summaries, we use an unsupervised evaluation method.
Following the recent success of using large language models in reference-free evaluation for text generation \citep{liu-etal-2023-g, chiang-lee-2023-large, shen-etal-2023-large}, we use GPT 3.5 to elicit rating for the generated summaries. We prompt the model using the reason-then-score prompt of \citet{shen-etal-2023-large}:\footnote{We use the \texttt{gpt-3.5-turbo-1106} model. This model is more recent than the one used in the evaluation of \citeauthor{shen-etal-2023-large}, but allows us to fit the entirety of the documents and summaries into the available tokens.}

\begin{quote}
    Score the following Summary given the corresponding Article with respect to relevance from one to five, where one indicates “irrelevance”, and five indicates “perfect relevance”. Note that relevance measures the Summary's selection of important content from the Article, whether the Summary grasps the main message of the Article without being overwhelmed by unnecessary or less significant details.
    
Article: \{article\}

Summary: \{summary\}

Provide your reason in one sentence, then give a final score:
\end{quote}

For each system, we evaluate all 683 summaries generated from the original documents which are used as templates for \balanced and \mono. For \balanced and \mono themselves we conserve resources and only evaluate summaries generated for two randomly selected inputs, resulting in 1366 ratings per system.

\section{Content Words}
\label{app:content_words}

\subsection{Motivation}

Our automatic template generation procedure only changes names and pronominal mentions, leaving content words unchanged. This can lead to unnatural occurrences, such as \textit{Chairman Diane Sasser}, when \textit{Chairwoman Diane Sasser} would be more appropriate. To check whether this is an issue in our experiments, we manually extend the automatically derived templates to also modify content words.

\begin{table}[t]
\centering
\begin{tabular}{|l|l|}
\hline
\textbf{Female} & \textbf{Male} \\ \hline
daughter             & son                  \\ \hline
mother               & father               \\ \hline
woman                & man                  \\ \hline
girl                 & boy                  \\ \hline
female               & male                 \\ \hline
sister               & brother              \\ \hline
daughters            & sons                 \\ \hline
mothers              & fathers              \\ \hline
women                & men                  \\ \hline
girls                & boys                 \\ \hline
females              & males                \\ \hline
sisters              & brothers             \\ \hline
aunt                 & uncle                \\ \hline
aunts                & uncles               \\ \hline
niece                & nephew               \\ \hline
nieces               & nephews              \\ \hline
wife                 & husband              \\ \hline
wives                & husbands             \\ \hline
actress              & actor                \\ \hline
actresses            & actors               \\ \hline
chairwoman           & chairman             \\ \hline
chairwomen           & chairmen             \\ \hline
mum                  & dad                  \\ \hline
mums                 & dads                 \\ \hline
waitress             & waiter               \\ \hline
waitresses           & waiters              \\ \hline
mistress             & lover                \\ \hline
\end{tabular}
\caption{Extended word list used to identify candidate articles for annotation.}
\label{tab:marker_words}
\end{table}

\begin{table*}
    \centering
    \begin{tabular}{|l|c|c|c|c|c|c|c|}
        \hline
        &\multicolumn{2}{c|}{BART} & \multicolumn{2}{c|}{Pegasus} & \multicolumn{3}{c|}{Llama-2 chat} \\
        & CNN & XSum & CNN & XSum & 7b & 13b & 70b \\
        \hline
Word List Inclusion & \renewcommand{\arraystretch}{0.6}\begin{tabular}{@{}c@{}} 0.00 \\\tiny s: 0.00,0.02 \\\tiny d: 0.00,0.05 \end{tabular} & \renewcommand{\arraystretch}{0.6}\begin{tabular}{@{}c@{}} 0.01 \\\tiny s: 0.00,0.04 \\\tiny d: 0.00,0.15 \end{tabular} & \renewcommand{\arraystretch}{0.6}\begin{tabular}{@{}c@{}} 0.02 \\\tiny s: 0.01,0.04 \\\tiny d: 0.00,0.08 \end{tabular} & \renewcommand{\arraystretch}{0.6}\begin{tabular}{@{}c@{}} 0.07 \\\tiny s: 0.03,0.10 \\\tiny d: 0.01,0.21 \end{tabular} & \renewcommand{\arraystretch}{0.6}\begin{tabular}{@{}c@{}} 0.08 \\\tiny s: 0.05,0.12 \\\tiny d: 0.04,0.14 \end{tabular} & \renewcommand{\arraystretch}{0.6}\begin{tabular}{@{}c@{}} 0.05 \\\tiny s: 0.03,0.06 \\\tiny d: 0.01,0.09 \end{tabular} & \renewcommand{\arraystretch}{0.6}\begin{tabular}{@{}c@{}} 0.06 \\\tiny s: 0.04,0.07 \\\tiny d: 0.02,0.09 \end{tabular}\\
Word List Inclusion (Orig.) & \renewcommand{\arraystretch}{0.6}\begin{tabular}{@{}c@{}} 0.05 \\\tiny s: 0.03,0.06 \\\tiny d: 0.00,0.11 \end{tabular} & \renewcommand{\arraystretch}{0.6}\begin{tabular}{@{}c@{}} 0.05 \\\tiny s: 0.02,0.09 \\\tiny d: 0.00,0.21 \end{tabular} & \renewcommand{\arraystretch}{0.6}\begin{tabular}{@{}c@{}} 0.06 \\\tiny s: 0.04,0.07 \\\tiny d: 0.00,0.14 \end{tabular} & \renewcommand{\arraystretch}{0.6}\begin{tabular}{@{}c@{}} 0.10 \\\tiny s: 0.07,0.13 \\\tiny d: 0.01,0.26 \end{tabular} & \renewcommand{\arraystretch}{0.6}\begin{tabular}{@{}c@{}} 0.06 \\\tiny s: 0.03,0.09 \\\tiny d: 0.01,0.11 \end{tabular} & \renewcommand{\arraystretch}{0.6}\begin{tabular}{@{}c@{}} 0.06 \\\tiny s: 0.05,0.08 \\\tiny d: 0.03,0.11 \end{tabular} & \renewcommand{\arraystretch}{0.6}\begin{tabular}{@{}c@{}} 0.04 \\\tiny s: 0.03,0.05 \\\tiny d: 0.00,0.09 \end{tabular}\\\hline
Entity Inclusion & \renewcommand{\arraystretch}{0.6}\begin{tabular}{@{}c@{}} 0.03 \\\tiny s: 0.00,0.07 \\\tiny d: 0.00,0.07 \end{tabular} & \renewcommand{\arraystretch}{0.6}\begin{tabular}{@{}c@{}} 0.05 \\\tiny s: 0.00,0.16 \\\tiny d: 0.00,0.23 \end{tabular} & \renewcommand{\arraystretch}{0.6}\begin{tabular}{@{}c@{}} 0.04 \\\tiny s: 0.01,0.08 \\\tiny d: 0.00,0.09 \end{tabular} & \renewcommand{\arraystretch}{0.6}\begin{tabular}{@{}c@{}} 0.01 \\\tiny s: 0.00,0.10 \\\tiny d: 0.00,0.26 \end{tabular} & \renewcommand{\arraystretch}{0.6}\begin{tabular}{@{}c@{}} 0.01 \\\tiny s: 0.00,0.08 \\\tiny d: 0.00,0.09 \end{tabular} & \renewcommand{\arraystretch}{0.6}\begin{tabular}{@{}c@{}} 0.06 \\\tiny s: 0.02,0.09 \\\tiny d: 0.01,0.10 \end{tabular} & \renewcommand{\arraystretch}{0.6}\begin{tabular}{@{}c@{}} 0.03 \\\tiny s: 0.00,0.06 \\\tiny d: 0.00,0.06 \end{tabular}\\
Entity Inclusion (Orig.) & \renewcommand{\arraystretch}{0.6}\begin{tabular}{@{}c@{}} 0.03 \\\tiny s: 0.00,0.07 \\\tiny d: 0.00,0.07 \end{tabular} & \renewcommand{\arraystretch}{0.6}\begin{tabular}{@{}c@{}} 0.05 \\\tiny s: 0.00,0.16 \\\tiny d: 0.00,0.24 \end{tabular} & \renewcommand{\arraystretch}{0.6}\begin{tabular}{@{}c@{}} 0.04 \\\tiny s: 0.01,0.08 \\\tiny d: 0.00,0.09 \end{tabular} & \renewcommand{\arraystretch}{0.6}\begin{tabular}{@{}c@{}} 0.01 \\\tiny s: 0.00,0.10 \\\tiny d: 0.00,0.26 \end{tabular} & \renewcommand{\arraystretch}{0.6}\begin{tabular}{@{}c@{}} 0.01 \\\tiny s: 0.00,0.08 \\\tiny d: 0.00,0.09 \end{tabular} & \renewcommand{\arraystretch}{0.6}\begin{tabular}{@{}c@{}} 0.06 \\\tiny s: 0.02,0.10 \\\tiny d: 0.01,0.10 \end{tabular} & \renewcommand{\arraystretch}{0.6}\begin{tabular}{@{}c@{}} 0.03 \\\tiny s: 0.00,0.06 \\\tiny d: 0.00,0.06 \end{tabular}\\\hline
Entity Hallucination & \renewcommand{\arraystretch}{0.6}\begin{tabular}{@{}c@{}} 0.30 \\\tiny s: 0.19,0.41 \\\tiny d: nan,nan \end{tabular} & \renewcommand{\arraystretch}{0.6}\begin{tabular}{@{}c@{}} 0.32 \\\tiny s: 0.29,0.34 \\\tiny d: 0.13,0.47 \end{tabular} & \renewcommand{\arraystretch}{0.6}\begin{tabular}{@{}c@{}} 0.45 \\\tiny s: 0.36,0.50 \\\tiny d: nan,nan \end{tabular} & \renewcommand{\arraystretch}{0.6}\begin{tabular}{@{}c@{}} 0.36 \\\tiny s: 0.32,0.39 \\\tiny d: 0.08,0.50 \end{tabular} & \renewcommand{\arraystretch}{0.6}\begin{tabular}{@{}c@{}} 0.35 \\\tiny s: 0.23,0.46 \\\tiny d: 0.22,0.45 \end{tabular} & \renewcommand{\arraystretch}{0.6}\begin{tabular}{@{}c@{}} 0.46 \\\tiny s: 0.40,0.50 \\\tiny d: 0.33,0.50 \end{tabular} & \renewcommand{\arraystretch}{0.6}\begin{tabular}{@{}c@{}} 0.39 \\\tiny s: 0.31,0.47 \\\tiny d: 0.14,0.50 \end{tabular}\\
Entity Hallucination (Orig.) & \renewcommand{\arraystretch}{0.6}\begin{tabular}{@{}c@{}} 0.26 \\\tiny s: 0.15,0.37 \\\tiny d: nan,nan \end{tabular} & \renewcommand{\arraystretch}{0.6}\begin{tabular}{@{}c@{}} 0.33 \\\tiny s: 0.31,0.35 \\\tiny d: 0.14,0.47 \end{tabular} & \renewcommand{\arraystretch}{0.6}\begin{tabular}{@{}c@{}} 0.50 \\\tiny s: 0.50,0.50 \\\tiny d: nan,nan \end{tabular} & \renewcommand{\arraystretch}{0.6}\begin{tabular}{@{}c@{}} 0.33 \\\tiny s: 0.29,0.36 \\\tiny d: 0.06,0.50 \end{tabular} & \renewcommand{\arraystretch}{0.6}\begin{tabular}{@{}c@{}} 0.17 \\\tiny s: 0.03,0.34 \\\tiny d: 0.02,0.39 \end{tabular} & \renewcommand{\arraystretch}{0.6}\begin{tabular}{@{}c@{}} 0.41 \\\tiny s: 0.33,0.47 \\\tiny d: 0.24,0.49 \end{tabular} & \renewcommand{\arraystretch}{0.6}\begin{tabular}{@{}c@{}} 0.40 \\\tiny s: 0.33,0.47 \\\tiny d: nan,nan \end{tabular}\\\hline
Distinguishability (Count) & \renewcommand{\arraystretch}{0.6}\begin{tabular}{@{}c@{}} 0.42 \\\tiny d: 0.35,0.49 \end{tabular} & \renewcommand{\arraystretch}{0.6}\begin{tabular}{@{}c@{}} 0.42 \\\tiny d: 0.34,0.50 \end{tabular} & \renewcommand{\arraystretch}{0.6}\begin{tabular}{@{}c@{}} 0.27 \\\tiny d: 0.19,0.35 \end{tabular} & \renewcommand{\arraystretch}{0.6}\begin{tabular}{@{}c@{}} 0.23 \\\tiny d: 0.15,0.31 \end{tabular} & \renewcommand{\arraystretch}{0.6}\begin{tabular}{@{}c@{}} 0.07 \\\tiny d: 0.02,0.11 \end{tabular} & \renewcommand{\arraystretch}{0.6}\begin{tabular}{@{}c@{}} 0.20 \\\tiny d: 0.11,0.28 \end{tabular} & \renewcommand{\arraystretch}{0.6}\begin{tabular}{@{}c@{}} 0.26 \\\tiny d: 0.19,0.33 \end{tabular}\\
Distinguishability (Count) (Orig.) & \renewcommand{\arraystretch}{0.6}\begin{tabular}{@{}c@{}} 0.27 \\\tiny d: 0.20,0.35 \end{tabular} & \renewcommand{\arraystretch}{0.6}\begin{tabular}{@{}c@{}} 0.32 \\\tiny d: 0.24,0.40 \end{tabular} & \renewcommand{\arraystretch}{0.6}\begin{tabular}{@{}c@{}} 0.19 \\\tiny d: 0.13,0.25 \end{tabular} & \renewcommand{\arraystretch}{0.6}\begin{tabular}{@{}c@{}} 0.17 \\\tiny d: 0.09,0.26 \end{tabular} & \renewcommand{\arraystretch}{0.6}\begin{tabular}{@{}c@{}} 0.12 \\\tiny d: 0.07,0.17 \end{tabular} & \renewcommand{\arraystretch}{0.6}\begin{tabular}{@{}c@{}} 0.15 \\\tiny d: 0.08,0.22 \end{tabular} & \renewcommand{\arraystretch}{0.6}\begin{tabular}{@{}c@{}} 0.10 \\\tiny d: 0.06,0.15 \end{tabular}\\\hline
Distinguishability (Dense) & \renewcommand{\arraystretch}{0.6}\begin{tabular}{@{}c@{}} 0.41 \\\tiny d: 0.34,0.49 \end{tabular} & \renewcommand{\arraystretch}{0.6}\begin{tabular}{@{}c@{}} 0.40 \\\tiny d: 0.31,0.47 \end{tabular} & \renewcommand{\arraystretch}{0.6}\begin{tabular}{@{}c@{}} 0.28 \\\tiny d: 0.21,0.36 \end{tabular} & \renewcommand{\arraystretch}{0.6}\begin{tabular}{@{}c@{}} 0.25 \\\tiny d: 0.18,0.32 \end{tabular} & \renewcommand{\arraystretch}{0.6}\begin{tabular}{@{}c@{}} 0.05 \\\tiny d: 0.00,0.10 \end{tabular} & \renewcommand{\arraystretch}{0.6}\begin{tabular}{@{}c@{}} 0.19 \\\tiny d: 0.12,0.25 \end{tabular} & \renewcommand{\arraystretch}{0.6}\begin{tabular}{@{}c@{}} 0.28 \\\tiny d: 0.21,0.34 \end{tabular}\\
Distinguishability (Dense) (Orig.) & \renewcommand{\arraystretch}{0.6}\begin{tabular}{@{}c@{}} 0.28 \\\tiny d: 0.21,0.35 \end{tabular} & \renewcommand{\arraystretch}{0.6}\begin{tabular}{@{}c@{}} 0.32 \\\tiny d: 0.24,0.40 \end{tabular} & \renewcommand{\arraystretch}{0.6}\begin{tabular}{@{}c@{}} 0.17 \\\tiny d: 0.10,0.25 \end{tabular} & \renewcommand{\arraystretch}{0.6}\begin{tabular}{@{}c@{}} 0.18 \\\tiny d: 0.10,0.25 \end{tabular} & \renewcommand{\arraystretch}{0.6}\begin{tabular}{@{}c@{}} 0.07 \\\tiny d: 0.02,0.11 \end{tabular} & \renewcommand{\arraystretch}{0.6}\begin{tabular}{@{}c@{}} 0.09 \\\tiny d: 0.01,0.16 \end{tabular} & \renewcommand{\arraystretch}{0.6}\begin{tabular}{@{}c@{}} 0.08 \\\tiny d: 0.03,0.14 \end{tabular}

\\\hline
\end{tabular}

    \caption{Results on our manually extended variants of \balanced and \mono for gender bias with content words altered to conform to entity gender. Since our annotations cover only a relatively small subset of the whole corpus, we also report the scores of summaries generated for the same inputs without content word modification for comparison (Orig.). We find that almost all scores fall within their respective confidence intervals.}
    \label{tab:results_content_words}
\end{table*}

\subsection{Annotation Procedure}

Since we found in preliminary experiments that many articles do not require any manual intervention, we first run an automatic filter over our dataset to identify candidate articles for annotation. We use an extended variant of our word list $W_g$ of \citet{liang_holistic_2022} reproduced in Table~\ref{tab:marker_words}. We then randomly sampled from these articles until we found 100 instances where at least one text span required manual intervention to adapt to entity gender.

During annotation, we identified text spans which should change in accordance with the gender of an entity in the document and which words should be used in either case (e.g. generating \textit{chairman} or \textit{chairwoman} depending on the gender of the entity occupying that position). We also considered the case where multiple entities might influence the realization of a particular word, like \textit{brothers}. In these cases, we also specify a neutral variant (e.g. siblings) to be used in case the referenced entities have different genders.

\subsection{Results}

We report results on gender bias with our modified inputs in Table~\ref{tab:results_content_words}. We find scores for modified inputs are very close to original scores when taking into account confidence intervals and exhibit the same trends. However, we note that the small number of inputs makes confidence intervals relatively wide.

\section{Intersectional Biases}

\label{app:ethnicity}

\begin{table*}
    \centering\begin{tabular}{|l|c|c|c|c|c|c|c|}
    \hline
&\multicolumn{2}{c|}{BART} & \multicolumn{2}{c|}{Pegasus} & \multicolumn{3}{c|}{Llama-2 chat} \\
Gender Assignment & CNN & XSum & CNN & XSum & 7b & 13b & 70b \\\hline

\multicolumn{7}{|l|}{Entity Inclusion Bias}\\
\hline

Random & \renewcommand{\arraystretch}{0.6}\begin{tabular}{@{}c@{}} 0.01 \\\tiny s: 0.00,0.03 \\\tiny d: 0.00,0.04 \end{tabular} & \renewcommand{\arraystretch}{0.6}\begin{tabular}{@{}c@{}} 0.17 \\\tiny s: 0.11,0.24 \\\tiny d: 0.08,0.29 \end{tabular} & \renewcommand{\arraystretch}{0.6}\begin{tabular}{@{}c@{}} 0.04 \\\tiny s: 0.00,0.09 \\\tiny d: 0.00,0.10 \end{tabular} & \renewcommand{\arraystretch}{0.6}\begin{tabular}{@{}c@{}} 0.02 \\\tiny s: 0.00,0.04 \\\tiny d: 0.00,0.05 \end{tabular} & \renewcommand{\arraystretch}{0.6}\begin{tabular}{@{}c@{}} 0.01 \\\tiny s: 0.00,0.04 \\\tiny d: 0.00,0.05 \end{tabular} & \renewcommand{\arraystretch}{0.6}\begin{tabular}{@{}c@{}} 0.01 \\\tiny s: 0.00,0.02 \\\tiny d: 0.00,0.03 \end{tabular} & \renewcommand{\arraystretch}{0.6}\begin{tabular}{@{}c@{}} 0.03 \\\tiny s: 0.01,0.05 \\\tiny d: 0.01,0.05 \end{tabular}\\
Black Male/White Female & \renewcommand{\arraystretch}{0.6}\begin{tabular}{@{}c@{}} 0.03 \\\tiny s: 0.01,0.04 \\\tiny d: 0.00,0.05 \end{tabular} & \renewcommand{\arraystretch}{0.6}\begin{tabular}{@{}c@{}} 0.19 \\\tiny s: 0.13,0.26 \\\tiny d: 0.10,0.31 \end{tabular} & \renewcommand{\arraystretch}{0.6}\begin{tabular}{@{}c@{}} 0.04 \\\tiny s: 0.00,0.09 \\\tiny d: 0.00,0.12 \end{tabular} & \renewcommand{\arraystretch}{0.6}\begin{tabular}{@{}c@{}} 0.02 \\\tiny s: 0.00,0.03 \\\tiny d: 0.00,0.04 \end{tabular} & \renewcommand{\arraystretch}{0.6}\begin{tabular}{@{}c@{}} 0.05 \\\tiny s: 0.02,0.08 \\\tiny d: 0.01,0.08 \end{tabular} & \renewcommand{\arraystretch}{0.6}\begin{tabular}{@{}c@{}} 0.02 \\\tiny s: 0.00,0.04 \\\tiny d: 0.00,0.04 \end{tabular} & \renewcommand{\arraystretch}{0.6}\begin{tabular}{@{}c@{}} 0.04 \\\tiny s: 0.02,0.06 \\\tiny d: 0.02,0.06 \end{tabular}\\
Black Male/White Male & \renewcommand{\arraystretch}{0.6}\begin{tabular}{@{}c@{}} 0.05 \\\tiny s: 0.03,0.07 \\\tiny d: 0.02,0.08 \end{tabular} & \renewcommand{\arraystretch}{0.6}\begin{tabular}{@{}c@{}} 0.11 \\\tiny s: 0.05,0.18 \\\tiny d: 0.03,0.21 \end{tabular} & \renewcommand{\arraystretch}{0.6}\begin{tabular}{@{}c@{}} 0.08 \\\tiny s: 0.03,0.13 \\\tiny d: 0.01,0.16 \end{tabular} & \renewcommand{\arraystretch}{0.6}\begin{tabular}{@{}c@{}} 0.05 \\\tiny s: 0.03,0.07 \\\tiny d: 0.03,0.08 \end{tabular} & \renewcommand{\arraystretch}{0.6}\begin{tabular}{@{}c@{}} 0.02 \\\tiny s: 0.00,0.05 \\\tiny d: 0.00,0.05 \end{tabular} & \renewcommand{\arraystretch}{0.6}\begin{tabular}{@{}c@{}} 0.01 \\\tiny s: 0.00,0.03 \\\tiny d: 0.00,0.03 \end{tabular} & \renewcommand{\arraystretch}{0.6}\begin{tabular}{@{}c@{}} 0.02 \\\tiny s: 0.00,0.04 \\\tiny d: 0.01,0.04 \end{tabular}\\
Black Female/White Male & \renewcommand{\arraystretch}{0.6}\begin{tabular}{@{}c@{}} 0.03 \\\tiny s: 0.01,0.05 \\\tiny d: 0.00,0.06 \end{tabular} & \renewcommand{\arraystretch}{0.6}\begin{tabular}{@{}c@{}} 0.24 \\\tiny s: 0.18,0.30 \\\tiny d: 0.13,0.38 \end{tabular} & \renewcommand{\arraystretch}{0.6}\begin{tabular}{@{}c@{}} 0.05 \\\tiny s: 0.00,0.10 \\\tiny d: 0.00,0.14 \end{tabular} & \renewcommand{\arraystretch}{0.6}\begin{tabular}{@{}c@{}} 0.01 \\\tiny s: 0.00,0.03 \\\tiny d: 0.00,0.04 \end{tabular} & \renewcommand{\arraystretch}{0.6}\begin{tabular}{@{}c@{}} 0.01 \\\tiny s: 0.00,0.04 \\\tiny d: 0.00,0.04 \end{tabular} & \renewcommand{\arraystretch}{0.6}\begin{tabular}{@{}c@{}} 0.01 \\\tiny s: 0.00,0.03 \\\tiny d: 0.00,0.04 \end{tabular} & \renewcommand{\arraystretch}{0.6}\begin{tabular}{@{}c@{}} 0.01 \\\tiny s: 0.00,0.03 \\\tiny d: 0.00,0.03 \end{tabular}\\
Black Female/White Female & \renewcommand{\arraystretch}{0.6}\begin{tabular}{@{}c@{}} 0.01 \\\tiny s: 0.00,0.03 \\\tiny d: 0.00,0.04 \end{tabular} & \renewcommand{\arraystretch}{0.6}\begin{tabular}{@{}c@{}} 0.12 \\\tiny s: 0.06,0.17 \\\tiny d: 0.02,0.23 \end{tabular} & \renewcommand{\arraystretch}{0.6}\begin{tabular}{@{}c@{}} 0.02 \\\tiny s: 0.00,0.07 \\\tiny d: 0.00,0.10 \end{tabular} & \renewcommand{\arraystretch}{0.6}\begin{tabular}{@{}c@{}} 0.04 \\\tiny s: 0.02,0.06 \\\tiny d: 0.01,0.07 \end{tabular} & \renewcommand{\arraystretch}{0.6}\begin{tabular}{@{}c@{}} 0.01 \\\tiny s: 0.00,0.04 \\\tiny d: 0.00,0.04 \end{tabular} & \renewcommand{\arraystretch}{0.6}\begin{tabular}{@{}c@{}} 0.01 \\\tiny s: 0.00,0.03 \\\tiny d: 0.00,0.03 \end{tabular} & \renewcommand{\arraystretch}{0.6}\begin{tabular}{@{}c@{}} 0.02 \\\tiny s: 0.00,0.04 \\\tiny d: 0.00,0.04 \end{tabular}\\

\hline
\multicolumn{7}{|l|}{Distinguishability (Count)}\\
\hline

Random & \renewcommand{\arraystretch}{0.6}\begin{tabular}{@{}c@{}} 0.19 \\\tiny d: 0.16,0.21 \end{tabular} & \renewcommand{\arraystretch}{0.6}\begin{tabular}{@{}c@{}} 0.23 \\\tiny d: 0.20,0.25 \end{tabular} & \renewcommand{\arraystretch}{0.6}\begin{tabular}{@{}c@{}} 0.16 \\\tiny d: 0.14,0.19 \end{tabular} & \renewcommand{\arraystretch}{0.6}\begin{tabular}{@{}c@{}} 0.10 \\\tiny d: 0.08,0.12 \end{tabular} & \renewcommand{\arraystretch}{0.6}\begin{tabular}{@{}c@{}} 0.01 \\\tiny d: -0.01,0.03 \end{tabular} & \renewcommand{\arraystretch}{0.6}\begin{tabular}{@{}c@{}} 0.04 \\\tiny d: 0.02,0.06 \end{tabular} & \renewcommand{\arraystretch}{0.6}\begin{tabular}{@{}c@{}} 0.01 \\\tiny d: -0.01,0.03 \end{tabular}\\
Black Male/White Female & \renewcommand{\arraystretch}{0.6}\begin{tabular}{@{}c@{}} 0.24 \\\tiny d: 0.22,0.26 \end{tabular} & \renewcommand{\arraystretch}{0.6}\begin{tabular}{@{}c@{}} 0.28 \\\tiny d: 0.25,0.31 \end{tabular} & \renewcommand{\arraystretch}{0.6}\begin{tabular}{@{}c@{}} 0.18 \\\tiny d: 0.16,0.21 \end{tabular} & \renewcommand{\arraystretch}{0.6}\begin{tabular}{@{}c@{}} 0.13 \\\tiny d: 0.11,0.16 \end{tabular} & \renewcommand{\arraystretch}{0.6}\begin{tabular}{@{}c@{}} 0.03 \\\tiny d: 0.01,0.05 \end{tabular} & \renewcommand{\arraystretch}{0.6}\begin{tabular}{@{}c@{}} 0.04 \\\tiny d: 0.02,0.06 \end{tabular} & \renewcommand{\arraystretch}{0.6}\begin{tabular}{@{}c@{}} 0.07 \\\tiny d: 0.05,0.09 \end{tabular}\\
Black Male/White Male & \renewcommand{\arraystretch}{0.6}\begin{tabular}{@{}c@{}} 0.20 \\\tiny d: 0.17,0.22 \end{tabular} & \renewcommand{\arraystretch}{0.6}\begin{tabular}{@{}c@{}} 0.25 \\\tiny d: 0.22,0.27 \end{tabular} & \renewcommand{\arraystretch}{0.6}\begin{tabular}{@{}c@{}} 0.15 \\\tiny d: 0.13,0.17 \end{tabular} & \renewcommand{\arraystretch}{0.6}\begin{tabular}{@{}c@{}} 0.09 \\\tiny d: 0.06,0.11 \end{tabular} & \renewcommand{\arraystretch}{0.6}\begin{tabular}{@{}c@{}} 0.02 \\\tiny d: -0.00,0.04 \end{tabular} & \renewcommand{\arraystretch}{0.6}\begin{tabular}{@{}c@{}} 0.02 \\\tiny d: 0.00,0.05 \end{tabular} & \renewcommand{\arraystretch}{0.6}\begin{tabular}{@{}c@{}} 0.04 \\\tiny d: 0.02,0.06 \end{tabular}\\
Black Female/White Male & \renewcommand{\arraystretch}{0.6}\begin{tabular}{@{}c@{}} 0.23 \\\tiny d: 0.21,0.26 \end{tabular} & \renewcommand{\arraystretch}{0.6}\begin{tabular}{@{}c@{}} 0.30 \\\tiny d: 0.27,0.33 \end{tabular} & \renewcommand{\arraystretch}{0.6}\begin{tabular}{@{}c@{}} 0.26 \\\tiny d: 0.24,0.29 \end{tabular} & \renewcommand{\arraystretch}{0.6}\begin{tabular}{@{}c@{}} 0.19 \\\tiny d: 0.16,0.21 \end{tabular} & \renewcommand{\arraystretch}{0.6}\begin{tabular}{@{}c@{}} 0.03 \\\tiny d: 0.01,0.05 \end{tabular} & \renewcommand{\arraystretch}{0.6}\begin{tabular}{@{}c@{}} 0.05 \\\tiny d: 0.03,0.07 \end{tabular} & \renewcommand{\arraystretch}{0.6}\begin{tabular}{@{}c@{}} 0.10 \\\tiny d: 0.08,0.12 \end{tabular}\\
Black Female/White Female & \renewcommand{\arraystretch}{0.6}\begin{tabular}{@{}c@{}} 0.21 \\\tiny d: 0.18,0.23 \end{tabular} & \renewcommand{\arraystretch}{0.6}\begin{tabular}{@{}c@{}} 0.24 \\\tiny d: 0.22,0.27 \end{tabular} & \renewcommand{\arraystretch}{0.6}\begin{tabular}{@{}c@{}} 0.22 \\\tiny d: 0.19,0.24 \end{tabular} & \renewcommand{\arraystretch}{0.6}\begin{tabular}{@{}c@{}} 0.12 \\\tiny d: 0.09,0.14 \end{tabular} & \renewcommand{\arraystretch}{0.6}\begin{tabular}{@{}c@{}} 0.01 \\\tiny d: -0.01,0.03 \end{tabular} & \renewcommand{\arraystretch}{0.6}\begin{tabular}{@{}c@{}} 0.01 \\\tiny d: -0.01,0.04 \end{tabular} & \renewcommand{\arraystretch}{0.6}\begin{tabular}{@{}c@{}} 0.05 \\\tiny d: 0.03,0.07 \end{tabular}\\

\hline
\multicolumn{7}{|l|}{Distinguishability (Dense)}\\
\hline

Random & \renewcommand{\arraystretch}{0.6}\begin{tabular}{@{}c@{}} 0.16 \\\tiny d: 0.13,0.19 \end{tabular} & \renewcommand{\arraystretch}{0.6}\begin{tabular}{@{}c@{}} 0.21 \\\tiny d: 0.19,0.24 \end{tabular} & \renewcommand{\arraystretch}{0.6}\begin{tabular}{@{}c@{}} 0.17 \\\tiny d: 0.14,0.19 \end{tabular} & \renewcommand{\arraystretch}{0.6}\begin{tabular}{@{}c@{}} 0.10 \\\tiny d: 0.08,0.13 \end{tabular} & \renewcommand{\arraystretch}{0.6}\begin{tabular}{@{}c@{}} 0.02 \\\tiny d: -0.00,0.04 \end{tabular} & \renewcommand{\arraystretch}{0.6}\begin{tabular}{@{}c@{}} 0.03 \\\tiny d: 0.01,0.05 \end{tabular} & \renewcommand{\arraystretch}{0.6}\begin{tabular}{@{}c@{}} 0.02 \\\tiny d: -0.00,0.04 \end{tabular}\\
Black Male/White Female & \renewcommand{\arraystretch}{0.6}\begin{tabular}{@{}c@{}} 0.24 \\\tiny d: 0.22,0.27 \end{tabular} & \renewcommand{\arraystretch}{0.6}\begin{tabular}{@{}c@{}} 0.28 \\\tiny d: 0.26,0.31 \end{tabular} & \renewcommand{\arraystretch}{0.6}\begin{tabular}{@{}c@{}} 0.19 \\\tiny d: 0.17,0.22 \end{tabular} & \renewcommand{\arraystretch}{0.6}\begin{tabular}{@{}c@{}} 0.13 \\\tiny d: 0.10,0.15 \end{tabular} & \renewcommand{\arraystretch}{0.6}\begin{tabular}{@{}c@{}} 0.01 \\\tiny d: -0.01,0.03 \end{tabular} & \renewcommand{\arraystretch}{0.6}\begin{tabular}{@{}c@{}} 0.03 \\\tiny d: 0.01,0.05 \end{tabular} & \renewcommand{\arraystretch}{0.6}\begin{tabular}{@{}c@{}} 0.05 \\\tiny d: 0.03,0.07 \end{tabular}\\
Black Male/White Male & \renewcommand{\arraystretch}{0.6}\begin{tabular}{@{}c@{}} 0.19 \\\tiny d: 0.17,0.22 \end{tabular} & \renewcommand{\arraystretch}{0.6}\begin{tabular}{@{}c@{}} 0.23 \\\tiny d: 0.21,0.26 \end{tabular} & \renewcommand{\arraystretch}{0.6}\begin{tabular}{@{}c@{}} 0.16 \\\tiny d: 0.13,0.18 \end{tabular} & \renewcommand{\arraystretch}{0.6}\begin{tabular}{@{}c@{}} 0.09 \\\tiny d: 0.07,0.12 \end{tabular} & \renewcommand{\arraystretch}{0.6}\begin{tabular}{@{}c@{}} 0.02 \\\tiny d: 0.00,0.04 \end{tabular} & \renewcommand{\arraystretch}{0.6}\begin{tabular}{@{}c@{}} 0.02 \\\tiny d: 0.00,0.04 \end{tabular} & \renewcommand{\arraystretch}{0.6}\begin{tabular}{@{}c@{}} 0.06 \\\tiny d: 0.04,0.08 \end{tabular}\\
Black Female/White Male & \renewcommand{\arraystretch}{0.6}\begin{tabular}{@{}c@{}} 0.24 \\\tiny d: 0.22,0.26 \end{tabular} & \renewcommand{\arraystretch}{0.6}\begin{tabular}{@{}c@{}} 0.29 \\\tiny d: 0.27,0.32 \end{tabular} & \renewcommand{\arraystretch}{0.6}\begin{tabular}{@{}c@{}} 0.26 \\\tiny d: 0.24,0.29 \end{tabular} & \renewcommand{\arraystretch}{0.6}\begin{tabular}{@{}c@{}} 0.18 \\\tiny d: 0.16,0.20 \end{tabular} & \renewcommand{\arraystretch}{0.6}\begin{tabular}{@{}c@{}} 0.04 \\\tiny d: 0.02,0.06 \end{tabular} & \renewcommand{\arraystretch}{0.6}\begin{tabular}{@{}c@{}} 0.05 \\\tiny d: 0.03,0.08 \end{tabular} & \renewcommand{\arraystretch}{0.6}\begin{tabular}{@{}c@{}} 0.09 \\\tiny d: 0.07,0.11 \end{tabular}\\
Black Female/White Female & \renewcommand{\arraystretch}{0.6}\begin{tabular}{@{}c@{}} 0.21 \\\tiny d: 0.19,0.24 \end{tabular} & \renewcommand{\arraystretch}{0.6}\begin{tabular}{@{}c@{}} 0.23 \\\tiny d: 0.21,0.26 \end{tabular} & \renewcommand{\arraystretch}{0.6}\begin{tabular}{@{}c@{}} 0.22 \\\tiny d: 0.19,0.25 \end{tabular} & \renewcommand{\arraystretch}{0.6}\begin{tabular}{@{}c@{}} 0.12 \\\tiny d: 0.10,0.15 \end{tabular} & \renewcommand{\arraystretch}{0.6}\begin{tabular}{@{}c@{}} 0.02 \\\tiny d: -0.00,0.04 \end{tabular} & \renewcommand{\arraystretch}{0.6}\begin{tabular}{@{}c@{}} 0.02 \\\tiny d: 0.00,0.04 \end{tabular} & \renewcommand{\arraystretch}{0.6}\begin{tabular}{@{}c@{}} 0.04 \\\tiny d: 0.02,0.06 \end{tabular}\\
\hline
\end{tabular}
    \caption{Bias scores for racial bias with black/white associated names with different gender assignments. \textit{Random} assigns gender uniformly at random, independently of race. This is the setting we report in the main paper.}
    \label{tab:results_race}
\end{table*}

While we have studied racial bias with randomly assigned gender in the main part of the paper, it is also interesting to consider the possibility of \textit{intersectional} effects that might become apparent when gender and race systematically correlate in the inputs.
We construct four additional datasets to study this, where we consider all four possible combinations of perceived entity race and binary gender.

Table~\ref{tab:results_race} shows that behavior is very similar between the random and intersectional settings. Interestingly, we find that for all models that have significantly non-zero distinguishability, it is highest when black and white coded entities are assigned opposite genders. Similarly, for BART XSum, inclusion bias is highest in these settings, although we note that none of the differences are significant. Overall we find no evidence of intersectional effects on our bias measures.

\section{Computational Infrastructure}

We ran most inference on a single node using four RX6800 GPUs. %
For Llama-2 13b and 70b we distributed additional experiments on two A100 and two H100 GPUs.

\end{document}